\PassOptionsToPackage{sort&compress,numbers}{natbib}
\documentclass{article}
\usepackage{amssymb}
\usepackage{amsmath, amsthm}
\usepackage{bm}
\usepackage{graphicx}
\usepackage{enumitem}
\usepackage{hyperref}
\usepackage{algorithm}
\usepackage{algorithmic}
\usepackage[rightcaption]{sidecap}
\usepackage{multirow} 
\usepackage{tcolorbox}
\usepackage{subcaption}
\usepackage{caption}
\usepackage{enumitem}
\setlist[itemize]{leftmargin=*, noitemsep}
\setlist[enumerate]{leftmargin=*, noitemsep}
\usepackage{booktabs}
\usepackage{listings}
\usepackage{graphicx} 
\usepackage{xcolor}
\usepackage{adjustbox} 
\usepackage{colortbl} 
\definecolor{lightgray}{gray}{0.9} 
\usepackage[T1]{fontenc}
\usepackage{lmodern}

\usepackage{fancybox}   
\usepackage{listings}
\usepackage[utf8]{inputenc}

\lstdefinelanguage{json}{
    basicstyle=\ttfamily\scriptsize,
    numbers=left,
    numberstyle=\tiny,
    stepnumber=1,
    numbersep=5pt,
    showstringspaces=false,
    breaklines=true,
    frame=single,
    backgroundcolor=\color{white},
    literate=
     *{0}{{\textcolor{blue}{0}}}{1}
      {1}{{\textcolor{blue}{1}}}{1}
      {2}{{\textcolor{blue}{2}}}{1}
      {3}{{\textcolor{blue}{3}}}{1}
      {4}{{\textcolor{blue}{4}}}{1}
      {5}{{\textcolor{blue}{5}}}{1}
      {6}{{\textcolor{blue}{6}}}{1}
      {7}{{\textcolor{blue}{7}}}{1}
      {8}{{\textcolor{blue}{8}}}{1}
      {9}{{\textcolor{blue}{9}}}{1}
      {:}{{\textcolor{red}{:}}}{1}
      {,}{{\textcolor{red}{,}}}{1}
      {\{}{{\textcolor{red}{\{}}}{1}
      {\}}{{\textcolor{red}{\}}}}{1}
      {[}{{\textcolor{red}{[}}}{1}
      {]}{{\textcolor{red}{]}}}{1},
}

\newcommand{\meanstd}[2]{#1\textcolor{gray}{\textsubscript{$\pm$#2}}}

\usepackage[preprint]{corl_2025} 

\title{\texttt{DriveMind}: A Dual Visual Language Model-based Reinforcement Learning Framework for Autonomous Driving}

\author{
  Dawood Wasif\thanks{This work has been submitted to the IEEE for possible publication (IEEE Transactions on Intelligent Vehicles). Copyright may be transferred without notice, after which this version may no longer be accessible.} \\
  Virginia Tech \\
  \texttt{dawoodwasif@vt.edu}
  \And
  Terrence J.\ Moore \\
  U.S.\ Army Research Laboratory \\
  \texttt{terrence.j.moore.civ@army.mil}
  \And
  Chandan K.\ Reddy \\
  Virginia Tech \\
  \texttt{reddy@cs.vt.edu}
  \And
  Frederica Free-Nelson \\
  U.S.\ Army Research Laboratory \\
  \texttt{frederica.f.nelson.civ@army.mil}
  \And
  Seunghyun Yoon \\
  KENTECH \\
  \texttt{syoon@kentech.ac.kr}
  \And
  Hyuk Lim \\
  KENTECH \\
  \texttt{hlim@kentech.ac.kr}
  \And
  Dan Dongseong Kim \\
  The University of Queensland \\
  \texttt{dan.kim@uq.edu.au}
  \And
  Jin-Hee Cho \\
  Virginia Tech \\
  \texttt{jicho@vt.edu}
}

\begin{document}
\maketitle



\begin{abstract}
End-to-end autonomous driving systems map sensor data directly to control commands, but remain opaque, lack interpretability, and offer no formal safety guarantees. While recent vision-language-guided reinforcement learning (RL) methods introduce semantic feedback, they often rely on static prompts and fixed objectives, limiting adaptability to dynamic driving scenes. We present \texttt{DriveMind}, a unified semantic reward framework that integrates: (i) a contrastive Vision-Language Model (VLM) encoder for stepwise semantic anchoring; (ii) a novelty-triggered VLM encoder-decoder, fine-tuned via chain-of-thought (CoT) distillation, for dynamic prompt generation upon semantic drift; (iii) a hierarchical safety module enforcing kinematic constraints (e.g., speed, lane centering, stability); and (iv) a compact predictive world model to reward alignment with anticipated ideal states. \texttt{DriveMind} achieves 19.4$\pm$2.3 km/h average speed, 0.98$\pm$0.03 route completion, and near-zero collisions in CARLA Town 2, outperforming baselines by over 4\% in success rate. Its semantic reward generalizes zero-shot to real dash-cam data with minimal distributional shift, demonstrating robust cross-domain alignment and potential for real-world deployment.
\end{abstract}

\keywords{Vision-Language Model; Reinforcement Learning; Autonomous Driving; Contrastive Semantic Reward; Chain-of-Thought Distillation}


\section{Introduction} \label{sec:introduction}

Recent advances in autonomous vehicles have shifted development from rigid pipelines to end-to-end neural policies mapping raw sensor streams directly to control commands \cite{chen2022learning, jia2023think, hu2023planning}. While these models offer streamlined architectures and strong benchmark performance, they raise critical deployment concerns. Their internal logic is opaque, complicating validation in safety-critical settings. They struggle to generalize to rare events like severe weather or infrastructure damage and lack formal guarantees on kinematic properties such as speed limits and lane-keeping. Further, they provide no natural interface for human oversight or explanation. These challenges motivate frameworks that combine deep network expressiveness with transparency, robustness, and provable safety.  

Meanwhile, Large Language Models (LLMs) and Vision Language Models (VLMs) have demonstrated human-level reasoning and visual grounding \cite{radford2021learning, xu2024drivegpt4, mao2023gpt}. Recent works like VLM-SR (Shaped Rewards) \cite{baumli2023vision}, VLM-RM (Reward Models) \cite{rocamonde2023vision}, and RoboCLIP (Language-Conditioned Robot Learning via Contrastive Language-Image Pretraining) \cite{sontakke2023roboclip} inject semantic feedback into Reinforcement Learning (RL), but rely on static prompts unsuited to evolving road conditions and overlook vehicle dynamics. Frequent VLM inference also incurs a high computational cost. As a result, current VLM-augmented systems trade off interpretability, adaptability, and real-time safety. To address this, we propose a framework that integrates dynamic VLM embeddings, hierarchical constraints, and a predictive world model to restore transparency and safety in autonomous driving.

We introduce \texttt{DriveMind}, which comprises four key modules, each targeting a key limitation of existing approaches. First, a contrastive VLM provides dense, context-aware semantic rewards by anchoring each bird’s-eye-view (BEV) frame against fixed “present” and “ideal” concepts, addressing the lack of informative feedback and poor context adaptation in traditional reward designs. Second, a novelty-triggered VLM encoder–decoder is invoked only when embedding drift exceeds a threshold, generating adaptive “present” (hazard) and “ideal” (goal) prompts to prevent reward hacking in repetitive or adversarial scenarios. This on‐demand prompting with asynchronous chain-of-thought (CoT) distillation using GPT-4 minimizes per‐step computation, refreshing embeddings only upon genuine scene changes, hence having negligible impact on the overall latency while maintaining efficiency.  Third, a hierarchical fusion module multiplicatively combines normalized kinematic factors, such as speed regulation, lane centering, heading alignment, lateral stability, to enforce a hard safety veto whenever any physical constraint is violated. Fourth, a compact world model forecasts the next semantic embedding and provides a predictive contrastive foresight reward, improving long-horizon credit assignment and anticipatory planning. Together, these modules yield an adaptive, interpretable, and provably safer reward framework for autonomous driving. Experiments on CARLA Town 2 demonstrate improved route completion, collision avoidance, and sample efficiency. Zero‐shot transfer to real dash-cam footage shows robust cross-domain alignment, confirming \texttt{DriveMind}’s promise for safer, more interpretable end‐to‐end driving.

This work makes the following \textbf{key contributions}:
\begin{itemize}
  \item We \emph{design} a dynamic dual-VLM architecture that extends static CLIP-based rewards~\cite{sontakke2023roboclip, radford2021learning} by anchoring each frame in a shared contrastive space and adds a novelty-triggered encoder–decoder to generate on-demand “present” and “ideal” prompts, eliminating the context insensitivity and reward hacking seen in prior fixed-prompt methods.
  \item We \emph{propose} a self-adjusting reward framework that integrates adaptive ideal-state contrastive signals, predictive foresight from a compact world model, and a hierarchical fusion of speed, lane centering, heading, and stability metrics, providing richer, scene-adaptive guidance compared to fixed-objective RL approaches.
  \item We \emph{validate} \texttt{DriveMind} via extensive experiments on CARLA Town 2 and zero-shot transfer to BDD100K dash-cam data, achieving $19.4 \pm 2.3$ km/h average speed, $0.98 \pm 0.03$ route completion, near-zero collision speed in simulation, and minimal distributional shift in real data (Wasserstein $= 0.028$, KS $= 0.105$), surpassing baselines on most safety and efficiency metrics.
\end{itemize}

\section{Related Work} \label{sec:related-work}
\textbf{End-to-End Autonomous Driving.} End-to-end autonomous driving has evolved from modular pipelines to unified neural policies mapping LiDAR and camera inputs to control commands. InterFuser~\cite{markhus2023reinforced} and UniAD~\cite{hu2023planning} pioneered Transformer and query-based architectures unifying perception, prediction, and planning. ThinkTwice~\cite{jia2023think} and ReasonNet~\cite{shao2023reasonnet} improved robustness under occlusion via iterative refinement and temporal context. TCP~\cite{wu2022trajectory} combined model-predictive control with learned forecasting, while LAV~\cite{chen2022learning} learned viewpoint-invariant latent spaces. Reinforcement-learning methods such as Latent DRL~\cite{tang2024efficient}, Roach~\cite{zhang2021end}, and skill-based approaches (ASAP-RL~\cite{wang2023efficient}, TaEcRL~\cite{zhou2023accelerating}) enhance training via demonstrations and primitives. However, end-to-end models remain opaque, complicating certification and offering no language-based explanation of behavior.

\textbf{Language Models in Driving Tasks.} LLMs' strengths in reasoning and dialogue have inspired integration into driving. DRIVEGPT4~\cite{xu2024drivegpt4} enables zero-shot visual query answering, but lacks direct control synthesis. GPT-Driver~\cite{mao2023gpt} tokenized road features for maneuver prediction, but suffered from discretization. LanguageMPC~\cite{wang2023empowering} grounded symbolic commands to continuous control via learned controllers, requiring careful prompting. LLM-Driver~\cite{chen2024driving} fused embeddings with language priors for traffic assessments without generating direct actions. While LLMs improve interpretability, their symbolic focus limits fine-grained, real-time control.

\textbf{Vision Language Models (VLMs) for Reward Design.} VLMs like CLIP~\cite{radford2021learning} have been used to automate reward shaping in RL. VLM-SR~\cite{baumli2023vision} binarizes frame-goal similarity, while RoboCLIP~\cite{sontakke2023roboclip} penalizes episodic semantic drift. VLM-RM~\cite{rocamonde2023vision} provides continuous rewards via text-difference projections but still uses static prompts. LORD~\cite{ye2024lord} penalizes undesired states without promoting positive behaviors. VLM-RL~\cite{huang2024vlm} contrasts dynamic embeddings for denser rewards, but struggles with evolving scenes and sparse feedback. These limitations motivate adaptive, context-aware reward mechanisms integrating dynamic semantics and predictive foresight.


\section{Proposed Approach: \texttt{DriveMind}}

\begin{figure*}
  \centering
  \includegraphics[width=0.9\textwidth]{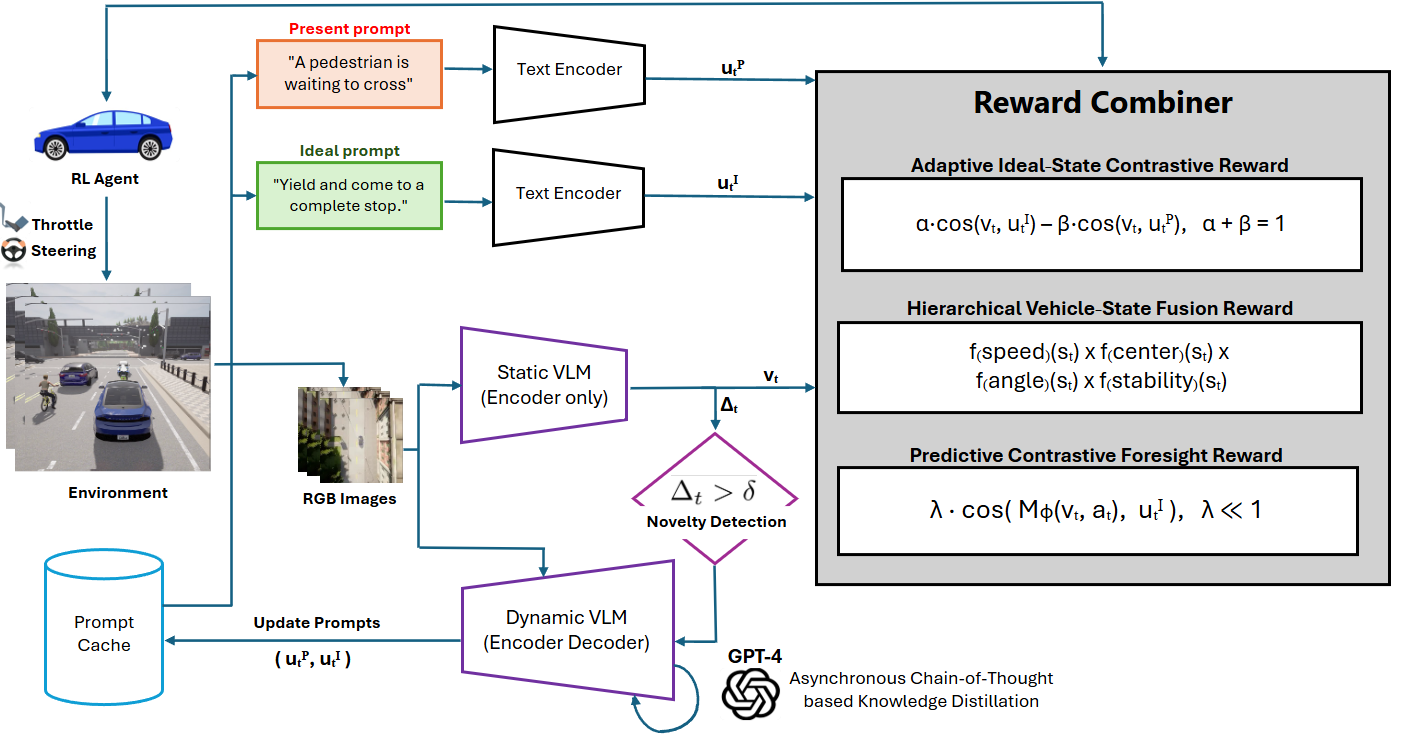}
    \caption{\textbf{Dual‑VLM architecture and reward pipeline of \texttt{DriveMind}:} A static contrastive VLM encoder maps each bird’s-eye-view image $\psi(s_t)$ to a fixed semantic embedding $v_t$, while a novelty detector asynchronously triggers a dynamic VLM encoder-decoder to update “present” and “ideal” prompt embeddings $(u^P_t,u^I_t)$ via distilled GPT‑4 chain-of-thought. These embeddings feed into the Reward Combiner, yielding (i) Adaptive Ideal-State Contrastive Reward, (ii) Hierarchical Vehicle-State Fusion Reward, and (iii) Predictive Contrastive Foresight Reward.}
  \label{fig:drivemind_architecture}
\end{figure*}

\subsection{Architectural Overview}

\texttt{DriveMind} integrates two VLMs, vehicle-state feedback, and a predictive world model into a self-adjusting reward framework for end-to-end autonomous driving. A frozen contrastive VLM encoder continuously encodes each BEV observation into a stable semantic embedding, while a novelty detector monitors deviations over a sliding window and triggers a VLM encoder-decoder, fine-tuned asynchronously via CoT distillation~\cite{deng2023implicit}, to generate context-specific “present” and “ideal” prompts as needed. Four normalized vehicle-state metrics, including speed regulation, lane centering, heading alignment, and lateral stability, are fused multiplicatively to enforce hard safety constraints, and a compact world model predicts the next embedding for one-step predictive foresight. These modules together produce the Adaptive Ideal-State Contrastive Reward (AICR) and Predictive Contrastive Foresight Module (PCFM), which, alongside classical task feedback, drive a Soft Actor-Critic agent toward safe, efficient, interpretable, and human-aligned behaviors (see Figure~\ref{fig:drivemind_architecture}).

\subsection{Static VLM Agent for Baseline Semantic Anchoring}

\texttt{DriveMind} employs a frozen contrastive VLM, $\mathrm{VLM}^C$, to provide a stable semantic reference at each timestep. Each BEV image $\psi(s_t)$ is embedded once by the image encoder and stored in a sliding window of past embeddings, while two text prompts stored in the prompt cache, “present” (e.g., “the car is about to crash”) and “ideal” (e.g., “no collisions occur”), are encoded once by the text encoder. These are combined to compute a per-step contrastive reward encouraging the current scene embedding $v_t$ to lie closer to the ideal than to the present concept:
\begin{equation}
r^{\mathrm{static}}_t = \tfrac{1}{2}\cos \bigl(v_t,u^I\bigr)\;-\;\tfrac{1}{2}\cos\bigl(v_t,u^P\bigr)\,,
\end{equation}
which, after normalization to $[0,1]$, acts as both an early training guide and a safety net against semantically unsafe regimes.

\subsection{Dynamic VLM Agent and Novelty-Triggered Prompt Caching}

\begin{figure*}
  \centering
  \includegraphics[width=\textwidth]{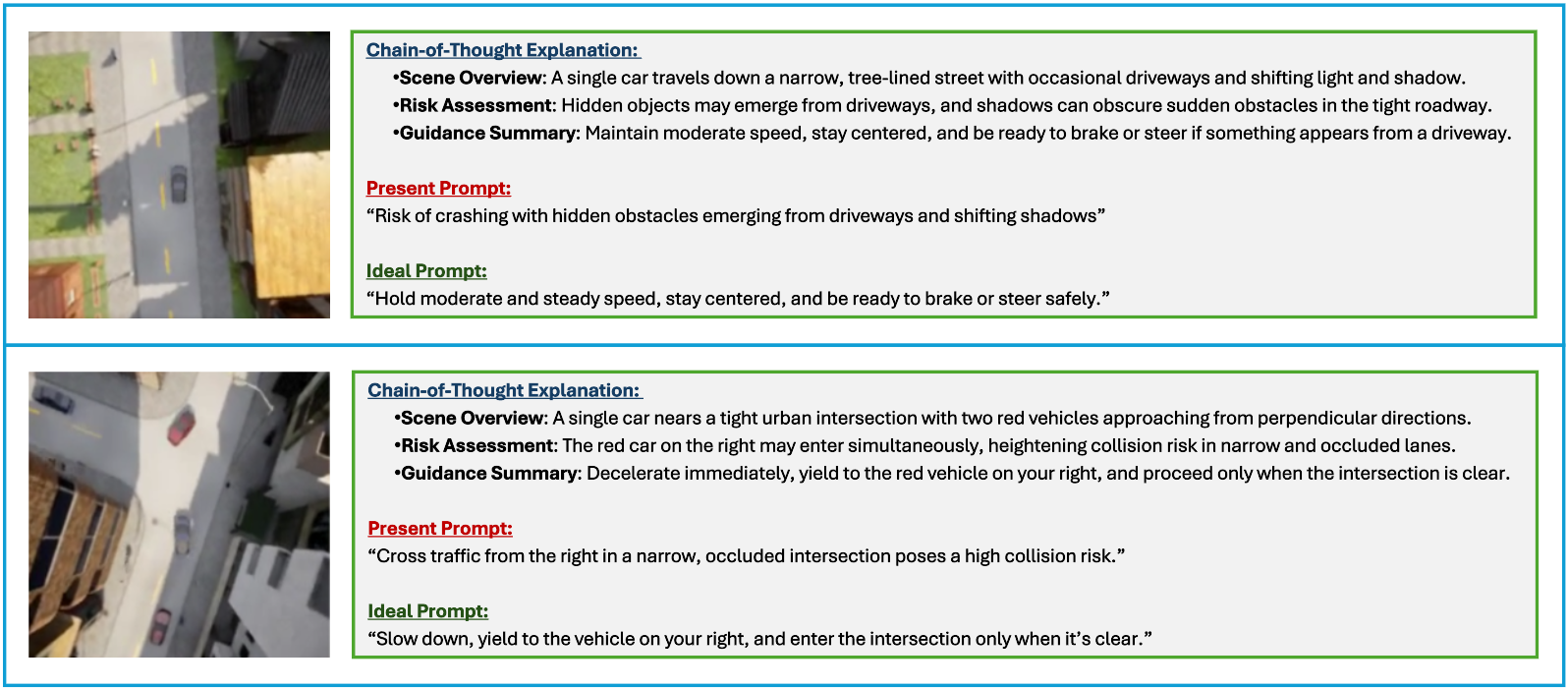}
    \caption{\textbf{Sample ground-truth labels from GPT-4 teacher for chain-of-thought distillation in dynamic VLM:} Examples from (i) a quiet residential street and (ii) a tight urban intersection. Each shows a bird’s-eye-view scene and the generated outputs: Scene Overview, Risk Assessment, Guidance Summary, and the present and ideal prompts used as distillation targets for producing the Adaptive ideal-State Contrastive Reward (AICR).}
  \label{fig:cot_kd_sample}
\end{figure*}

To capture rare or evolving driving scenarios (examples shown in Figure~\ref{fig:cot_kd_sample}), \texttt{DriveMind} maintains a cache $\mathcal{B}$ of recent static embeddings and computes a novelty score
\begin{equation}
\Delta_t = \min_{v' \in \mathcal{B}} \|v_t - v'\|_2\,,
\end{equation}
invoking a lightweight encoder-decoder VLM, $\mathrm{VLM}^D$, only when $\Delta_t$ exceeds a threshold $\delta$. Upon triggering, context-specific “present” and “ideal” prompts are generated, re-embedded, stored in the prompt-cache (evicting the oldest entry), and reused until further scene changes, balancing semantic adaptability with real-time efficiency.
   
\subsection{Reward Components}

\subsubsection{Adaptive Contrastive Semantic Reward}
\texttt{DriveMind}'s \textit{Adaptive Contrastive Semantic Reward} sculpts a dense, context-specific signal at each timestep using dynamic ``present'' and ``ideal'' prompt embeddings. A lightweight encoder-decoder VLM, fine-tuned via CoT distillation using GPT-4 teacher, generates language descriptions of current hazards (``present'') and desired outcomes (``ideal'') only when the scene significantly deviates from past experience. These prompts are re-embedded into the static contrastive space, enabling \texttt{DriveMind} to measure alignment between the current visual embedding and each context-specific concept. The reward combines a positive term for proximity to the ``ideal'' embedding and a negative term for proximity to the ``present'' embedding, simultaneously encouraging safe progress and discouraging hazardous states. Clipping and normalization ensure numerical stability, while adjustable positive and negative weights balance proactive safety and reactive hazard avoidance, yielding nuanced, adaptive guidance beyond static text prompts.

\subsubsection{Hierarchical Vehicle-State Fusion Reward}
While semantic objectives define desired outcomes, real-world driving demands strict kinematic safety. \texttt{DriveMind}'s \textit{Hierarchical Vehicle-State Synthesis Reward} enforces this by fusing four normalized metrics, including speed regulation, lane centering, heading alignment, and lateral stability, into a single multiplicative term. Each factor is independently scaled between zero (complete violation) and one (ideal execution): the speed component penalizes deviations from context-appropriate target velocities (e.g., smooth deceleration in a school zone), the centering component measures lateral offset from the lane midline, the heading component captures angular misalignment with road direction, and the stability component penalizes abrupt lateral oscillations or yaw spikes. Multiplicative fusion enforces a strict safety veto: if any kinematic constraint is violated (i.e., its normalized score is zero), the entire reward term collapses. This implements a logical “AND,” ensuring no positive reward unless \textit{all} safety-critical factors are satisfied.

\subsubsection{Predictive Contrastive Foresight Module}
To improve temporal credit assignment for maneuvers requiring lookahead, such as smoothing lane changes or planning gentle deceleration, \texttt{DriveMind}'s \textit{Predictive Contrastive Foresight Module} trains a compact world model to forecast the next-step visual embedding from the current embedding and action. This prediction is evaluated against the ``ideal'' prompt embedding using the same cosine-based contrastive metric as the semantic reward. By modestly rewarding actions that bring the predicted embedding closer to the ideal state, \texttt{DriveMind} provides foresight into the long-term semantic consequences of each decision. Proper weighting ensures this predictive signal complements rather than overwhelms immediate semantic and safety rewards, enabling smoother, more anticipatory driving that better balances efficiency, comfort, and safety over extended trajectories.

\noindent
Combining all components yields the final per‐step reward:
\begin{equation}
\label{eq:combined_reward}
\begin{aligned}
r_t &= r_{\mathrm{task},t}
  + \underbrace{\rho_1\bigl(r_{\mathrm{speed}}(s_t)\,f_{\mathrm{center}}(s_t)\,f_{\mathrm{angle}}(s_t)\,f_{\mathrm{stability}}(s_t)\bigr)}_{\substack{\text{Hierarchical Vehicle-State}\\\text{Fusion Reward}}}\\ 
    &\quad
  + \underbrace{\rho_2\bigl(\alpha\,\cos(v_t,u^I_t)\;-\;\beta\,\cos(v_t,u^P_t)\bigr)}_{\substack{\text{Adaptive Ideal-State}\\\text{Contrastive Reward}}}
  + \underbrace{\lambda\,\cos\bigl(M_{\phi}(v_t,a_t),\,u^I_t\bigr)}_{{\substack{\text{Predictive Contrastive}\\\text{Foresight Reward}}}}\,.
\end{aligned}
\end{equation}
Here, $r_{\mathrm{task},t}$ is the task-specific reward at time $t$; $\rho_1$ and $\rho_2$ are scaling factors for the safety synthesis and semantic contrastive terms; $r_{\mathrm{speed}}(s_t)$, $f_{\mathrm{center}}(s_t)$, $f_{\mathrm{angle}}(s_t)$, and $f_{\mathrm{stability}}(s_t)$ are normalized metrics for speed regulation, lane centering, heading alignment, and lateral stability at state $s_t$; $\alpha$ and $\beta$  (with $\alpha,\beta=\tfrac{1}{2}$) weight the contrastive reward between the current VLM embedding $v_t$ and the dynamic ``ideal'' and ``present'' prompt embeddings $u^I_t$ and $u^P_t$; $M_{\phi}(v_t,a_t)$ is the world model’s one-step predicted embedding given the current embedding and action; and $\lambda$ balances the predictive foresight term based on its cosine similarity to $u^I_t$. 


\section{Experiments and Results}
We design our experiments to address the following questions:
\begin{itemize}
    \item[] \textbf{Q1:} How does \texttt{DriveMind} impact driving performance (e.g., safety, efficiency, and comfort) compared to state-of-the-art reward learning methods in the CARLA simulator?
    \item[] \textbf{Q2:} What are the contributions of each component in \texttt{DriveMind} to overall performance?
    \item[] \textbf{Q3:} Can \texttt{DriveMind}'s semantic rewards generalize zero-shot to real dash-cam data?
\end{itemize}

\subsection{Experimental Setup}
\textbf{Dataset \& Environments.} \texttt{DriveMind} is evaluated first in the CARLA simulator and then on real-world dash-cam imagery to assess both controlled performance and domain transfer. For simulation benchmarks, we train exclusively on CARLA Town 2. During training, we sample driving scenarios by uniformly selecting two distinct spawn points from a predefined waypoint set and computing the shortest path via $A^*$; routes exceeding 3 $km$ are rejected and resampled to ensure bounded episode lengths. To capture varying traffic conditions, we deploy three density levels: “empty” (no other vehicles), “regular” (20 vehicles), and “dense” (40 vehicles). Each episode terminates upon collision, lane departure beyond 3 $m$, or being stuck (speed below 1 $km/h$ for 90 seconds). For evaluation, we freeze all vision, language, and RL components, and test on four held-out maps (Towns 1, 3, 4, and 5) under the same three densities, reporting average speed, route completion, distance traveled, collision speed, and success rate.

To measure real-world generalization, we leverage 10,000 bird’s-eye-view projections from the BDD100K validation split. Each dash-cam frame is converted into a bird's-eye-view semantic segmentation image using a pretrained road-and-vehicle mask model at $224 \times 224$ resolution. The policy executes a single control step using this image and the resulting “ideal” and “present” reward scores are recorded. We then compute the distribution of the \textit{Adaptive Ideal-State Contrastive Reward} (AICR) across these frames and compare it to the CARLA Town 2 reference using the Wasserstein-1 distance and Kolmogorov–Smirnov statistic.

\textbf{Observations \& Actions.} At each timestep, the agent receives an observation $o_t$ consisting of three parts. First, a bird’s-eye-view semantic image, resized to $224 \times 224$ px, fuses camera and LiDAR segmentation of lanes, vehicles, pedestrians, and obstacles. Second, an ego-state vector $e_t = (v_t, \theta_t, \omega_t)$ captures normalized longitudinal speed, steering angle, and yaw rate from onboard sensors. Third, a sequence of $K=10$ future waypoints $(x_{t+1}, y_{t+1}, \dots, x_{t+K}, y_{t+K})$, sampled every 5 m in the vehicle’s local frame and scaled to $[-1,1]$, provides route guidance. Together, these inputs form the full policy observation. The agent outputs a continuous action $a_t = (a^{(1)}, a^{(2)}) \in [-1,1]^2$, where $a^{(1)}$ maps linearly to steering angle and $a^{(2)}$ splits into throttle (positive) and brake (negative), each normalized to $[0,1]$. This compact interface enables real-time, smooth control in response to learned reward signals.

\textbf{Networks \& Learning.}
\texttt{DriveMind} is trained using a custom Soft Actor-Critic (SAC)~\cite{haarnoja2018soft} implementation inspired by Stable-Baselines3. The actor and critic networks share a lightweight feature extractor that fuses a compact convolutional encoder for bird’s-eye-view (BEV) images with separate MLP branches for the 3D ego-state vector and 20D waypoint input. Semantic grounding is provided by two VLMs: a frozen CLIP ViT-bigG-14 encoder, used for reward shaping via alignment with language prompts, and a dynamic SmolVLM-256M encoder-decoder, fine-tuned online via GPT-4o-mini through structured CoT distillation. When triggered, SmolVLM generates ``present" and ``ideal" prompts along with a natural language CoT explanation in JSON format.

SAC uses a replay buffer of size $10^5$, samples mini-batches of 256 transitions, and performs one gradient update per environment step. Actor and Critic networks are optimized using Adam with a learning rate of $3\times10^{-4}$, automatic entropy tuning, and Polyak averaging ($\tau=5\times10^{-3}$) for target networks. SmolVLM fine-tuning is performed asynchronously using Low-Rank Adaptation (LoRA) with one-step updates per sample, using AdamW with a learning rate of $5\times10^{-5}$. Only LoRA adapter parameters are trained, while the base model and image encoder remain frozen. All experiments use mixed-precision (i.e., 16-bit floating point, FP16), and random seeds are fixed across environments and modules. Hyperparameters and network settings are constant across all baselines to ensure fair comparison over $10^6$ environment interactions.

\textbf{Metrics.} To evaluate both efficiency and safety in CARLA, we use five scalar metrics~\cite{dosovitskiy2017carla}:  
(1) \textbf{\em Average Speed (AS)} is the mean forward velocity per episode, indicating how quickly the agent completes its route; (2) \textbf{\em Route Completion (RC)} is the fraction of episodes where the vehicle reaches the goal without early termination, reflecting reliability under traffic variation; (3) \textbf{\em Total Distance (TD)} is the cumulative path length per episode, capturing overall coverage and exploration;  (4) \textbf{\em Collision Speed (CS)} is the mean speed at collision time, serving as a proxy for impact severity; and (5) \textbf{\em Success Rate (SR)} is the percentage of episodes that reach the goal without collisions.

For zero-shot transfer to real dash-cam data, we analyze the AICR across 10,000 frames and compare its empirical distribution to simulation using the Wasserstein-1 distance \cite{villani2008optimal} and Kolmogorov–Smirnov statistic \cite{smirnov1939estimation}, quantifying the domain gap in semantic reward alignment.

\textbf{Baselines.} We compare \texttt{DriveMind} against fourteen state-of-the-art reward methods across three categories. In \textbf{\em expert-designed rewards}, Trajectory-based Inverse Reinforcement Learning (TIRL) SAC~\cite{cao2022trustworthy} and TIRL PPO~\cite{cao2022trustworthy} apply a binary collision penalty only, while Chen SAC~\cite{chen2021interpretable} balances collision avoidance, speed, lane keeping, and steering smoothness. Adaptive Skills, Adaptive Partitions (ASAP)-RL PPO~\cite{wang2023efficient} incentivizes progress and overtaking alongside safety, and ChatScene SAC~\cite{zhang2024chatscene} and ChatScene PPO~\cite{zhang2024chatscene} promote smoothness with constant learning signals. In \textbf{\em LLM-designed rewards}, Revolve~\cite{hazra2024revolve} generates candidate reward functions via LLM-driven code search, while Revolve-Auto~\cite{hazra2024revolve} iteratively refines them using self-labeled trajectories \cite{hazra2024revolve}. In \textbf{\em VLM-designed rewards}, methods use fixed or paired language prompts: VLM-SR \cite{baumli2023vision}, RoboCLIP \cite{sontakke2023roboclip}, VLM-RM \cite{rocamonde2023vision}, Large Models-based Opposite Reward Design (LORD) and LORD-Speed \cite{ye2024lord}, and VLM-RL \cite{huang2024vlm}. All baselines share \texttt{DriveMind}’s network architectures and hyperparameters to ensure a fair comparison.
\begin{table}[ht]
\centering
\caption{\centering Performance Comparison of \texttt{DriveMind} Against Expert-Designed, LLM-Based, and VLM-Based Reward Methods During Testing. Reported Values Are Mean $\pm$ Standard Deviation Over Three Seeds. Best Results Are \textbf{Bolded}.}
\label{tab:test_results}
\begin{adjustbox}{width=1.0\textwidth}

\begin{tabular}{llccccc}
\toprule
Category & Model & AS $\uparrow$ & RC $\uparrow$ & TD $\uparrow$ & CS $\downarrow$ & SR $\uparrow$ \\
\midrule
\multirow{2}{*}{Expert-Designed (Binary)} 
    & TIRL-SAC \cite{cao2022trustworthy}       & \meanstd{0.37}{0.28} & \meanstd{0.01}{0.001} & \meanstd{4.7}{3.5}   & \meanstd{0.24}{0.34} & \meanstd{0.00}{0.00} \\
    & TIRL-PPO \cite{cao2022trustworthy}       & \meanstd{0.43}{0.23} & \meanstd{0.01}{0.005} & \meanstd{14.8}{9.7}  & \meanstd{0.10}{0.15} & \meanstd{0.00}{0.00} \\
\midrule
\multirow{4}{*}{Expert-Designed (Summation)} 
    & Chen-SAC \cite{chen2021interpretable}      & \meanstd{21.4}{1.16} & \meanstd{0.29}{0.12}  & \meanstd{663.6}{286.7} & \meanstd{2.07}{2.21} & \meanstd{0.08}{0.08} \\
    & ASAP-RL-PPO \cite{wang2023efficient}   & \meanstd{1.25}{0.30} & \meanstd{0.01}{0.00}  & \meanstd{28.1}{3.4}   & \meanstd{0.61}{0.61} & \meanstd{0.00}{0.00} \\
    & ChatScene-SAC \cite{zhang2024chatscene}  & \meanstd{17.7}{0.12} & \meanstd{0.88}{0.03}  & \meanstd{1763.2}{90.9} & \meanstd{1.18}{0.46} & \meanstd{0.73}{0.05} \\
    & ChatScene-PPO \cite{zhang2024chatscene}  & \meanstd{15.3}{0.33} & \meanstd{0.78}{0.05}  & \meanstd{1515.6}{129.1} & \meanstd{0.89}{0.32} & \meanstd{0.63}{0.05} \\
\midrule
\multirow{2}{*}{LLM-Based}
    & Revolve \cite{hazra2024revolve}       & \meanstd{18.4}{0.03} & \meanstd{0.92}{0.11}  & \meanstd{1915.3}{248.3} & \meanstd{1.53}{2.16} & \meanstd{0.83}{0.24} \\
    & Revolve-Auto \cite{hazra2024revolve}   & \meanstd{17.2}{0.76} & \meanstd{0.80}{0.06}  & \meanstd{1539.6}{147.5} & \meanstd{1.65}{0.28} & \meanstd{0.63}{0.05} \\
\midrule
\multirow{6}{*}{VLM-Based}
    & VLM-SR \cite{baumli2023vision}       & \meanstd{0.53}{0.27} & \meanstd{0.02}{0.00}  & \meanstd{47.9}{9.2}   & \meanstd{0.18}{0.25} & \meanstd{0.00}{0.00} \\
    & RoboCLIP \cite{sontakke2023roboclip}       & \meanstd{0.44}{0.05} & \meanstd{0.07}{0.03}  & \meanstd{146.3}{62.3}  & \meanstd{1.05}{0.58} & \meanstd{0.00}{0.00} \\
    & VLM-RM \cite{rocamonde2023vision}         & \meanstd{0.20}{0.05} & \meanstd{0.02}{0.01}  & \meanstd{35.9}{25.8}   & \textbf{\meanstd{0.003}{0.005}} & \meanstd{0.00}{0.00} \\
    & LORD \cite{ye2024lord}           & \meanstd{0.17}{0.08} & \meanstd{0.02}{0.02}  & \meanstd{45.1}{57.1}   & \meanstd{0.02}{0.02} & \meanstd{0.00}{0.00} \\
    & LORD-Speed \cite{ye2024lord}     & \meanstd{18.9}{0.36} & \meanstd{0.87}{0.05}  & \meanstd{1783.4}{172.8} & \meanstd{2.80}{1.16} & \meanstd{0.67}{0.05} \\
    & VLM-RL \cite{huang2024vlm}         & \meanstd{19.3}{1.29} & \meanstd{0.97}{0.03}  & \meanstd{2028.2}{96.60} & \meanstd{0.02}{0.03} & \meanstd{0.93}{0.04} \\
\midrule
\rowcolor{lightgray}
\textbf{Ours}
    & \texttt{DriveMind}     & \textbf{\meanstd{19.4}{2.34}} & \textbf{\meanstd{0.98}{0.03}} & \textbf{\meanstd{2083.1}{50.09}} & {\meanstd{0.01}{0.07}} & \textbf{\meanstd{0.97}{0.06}} \\
\bottomrule
\end{tabular}
\end{adjustbox}
\end{table}

To assess real-time deployability, we profile runtime latency for key components, which is detailed in the Appendix~\ref{appen:real-time-deloyability}. 

\subsection{Performance Comparison on CARLA}

Table~\ref{tab:test_results} reports the mean $\pm$ standard deviation performance over three seeds for \texttt{DriveMind} and fourteen baselines on CARLA Town 2. \texttt{DriveMind} achieves an average speed of \textbf{19.4 $\pm$ 2.34 km/h}, a route completion rate of \textbf{0.98 $\pm$ 0.03}, and travels \textbf{2083.1 $\pm$ 50.09 m} per episode, while maintaining a near-zero collision speed (0.01 $\pm$ 0.07 km/h) and a success rate of \textbf{0.97 $\pm$ 0.06}. In contrast, purely binary penalties (TIRL-SAC/PPO) fail to produce meaningful motion~\cite{cao2022trustworthy}, and hand-tuned summation rewards such as Chen-SAC achieve moderate speeds but incur significant collisions and reduced success rates~\cite{chen2021interpretable}. ASAP-RL-PPO drives aggressively yet unsafely~\cite{wang2023efficient}, while ChatScene variants prioritize smoothness at the expense of task completion~\cite{zhang2024chatscene}. LLM-generated rewards (Revolve, Revolve-Auto) achieve high route completion but introduce unpredictable safety trade-offs~\cite{hazra2024revolve}. VLM-based methods (VLM-SR, RoboCLIP, VLM-RM, LORD) either stagnate or learn inadequate steering behaviors~\cite{baumli2023vision}, and even VLM-RL, which combines positive and negative prompts, exhibits nonzero collision speeds~\cite{huang2024vlm}. In contrast, \texttt{DriveMind}’s integration of adaptive contrastive semantics, hierarchical safety fusion, and predictive foresight yields state-of-the-art efficiency alongside near-perfect safety, validating its unified reward design.

\subsection{Ablation Studies}

Table~\ref{tab:ablation_modules} summarizes the impact of removing each core \texttt{DriveMind} module, adaptive contrastive reward (NoAICR), hierarchical vehicle-state fusion reward (NoHVFR), and predictive foresight (NoPCFM), on CARLA Town 2 performance. Removing hierarchical fusion collapses driving ability (AS 0.58 km/h, RC 0.02, SR 0.00), underscoring its role as a critical safety veto. Omitting the contrastive reward leads to slower, less reliable driving (AS 17.6 km/h, RC 0.82, SR 0.67) with elevated collision speeds (CS 1.66 km/h), highlighting its importance for semantic guidance. In contrast, removing predictive foresight only slightly reduces route completion and success rates (RC 0.92, SR 0.87), indicating that foresight refines but is not critical to core safety and efficiency. The full \texttt{DriveMind} (AS 19.4 km/h, RC 0.98, CS 0.01 km/h, SR 0.97) clearly outperforms all ablated variants, validating the synergy of all components.


\begin{table}[ht]
  \centering
  \caption{\centering Ablation of Key \texttt{DriveMind} Modules: Mean $\pm$ Standard Deviation Over Three Seeds}
  \label{tab:ablation_modules}
  \begin{tabular}{lcccc}
    \toprule
    Variant            & AS $\uparrow$      & RC $\uparrow$      & CS $\downarrow$       & SR $\uparrow$       \\
    \midrule
    DriveMind, NoAICR   & $17.6\pm0.95$       & $0.82\pm0.07$      & $1.66\pm0.45$         & $0.67\pm0.13$       \\
    DriveMind, NoHVFR   & $0.58\pm0.25$       & $0.02\pm0.00$      & $0.14\pm0.21$         & $0.00\pm0.00$       \\
    DriveMind, NoPCFM   & $19.2\pm1.14$       & $0.92\pm0.05$      & $0.02\pm0.02$         & $0.87\pm0.06$       \\
    \rowcolor{lightgray}
    DriveMind, Full     & $\mathbf{19.4\pm2.34}$ & $\mathbf{0.98\pm0.03}$ & $\mathbf{0.01\pm0.07}$   & $\mathbf{0.97\pm0.06}$ \\
    \bottomrule
  \end{tabular}
\end{table}

\subsection{Real-World Generalization}

Table~\ref{tab:real_world_generalization} compares the Adaptive Ideal-State Contrastive Reward (AICR) distributions in CARLA Town 2 versus 10,000 BDD100K bird's-eye-view frames. On real data, the mean reward drops slightly from 0.130 to 0.118 and the standard deviation rises modestly from 0.080 to 0.085. The fraction of positive‐reward frames falls from 88.0 \% to 77.4 \%, while negative‐reward frames increase from 12.0 \% to 22.6 \%. Crucially, the Wasserstein‐1 distance (0.028) and the Kolmogorov-Smirnov statistic (0.105) remain low, indicating that the two empirical distributions closely align. These results confirm that \texttt{DriveMind}’s learned semantic objectives transfer effectively, maintaining robust AICR behavior in unconstrained, real‐world driving scenes without any additional fine‐tuning.  

\definecolor{rowgray}{gray}{0.95}
\begin{table}[ht]
\centering
\caption{\centering Real-World Generalization of \texttt{DriveMind}’s AICR Reward: Comparison Between CARLA Town 2 and BDD100K Dash-Cam Frames}
\label{tab:real_world_generalization}
\begin{adjustbox}{width=\textwidth}
\begin{tabular}{lcccccc}
\toprule
Dataset & Mean $r_{\rm AICR}$ & STD $r_{\rm AICR}$ & \% $r_{\rm AICR} > 0$ & \% $r_{\rm AICR} < 0$ & EMD (CARLA $\rightarrow$ Real) & KS-stat \\
\midrule
CARLA Town 2 & 0.130 & 0.080 & 88.0\% & 12.0\% & ,  & ,  \\
\rowcolor{rowgray}
BDD100K (10K frames) & 0.118 & 0.085 & 77.4\% & 22.6\% & 0.028 & 0.105 \\
\bottomrule
\end{tabular}
\end{adjustbox}
\end{table}


\section{Conclusions \& Future Work}
\label{sec:conclusions}

\textbf{Summary.} We introduced \texttt{DriveMind}, a unified reward framework combining: (i) a frozen contrastive VLM for per-step semantic anchoring, (ii) a novelty-triggered encoder-decoder VLM fine-tuned via CoT distillation for dynamic “present” and “ideal” prompts, (iii) hierarchical fusion of kinematic metrics enforcing hard safety constraints, and (iv) a compact world model for predictive contrastive foresight. Integrated into adaptive contrastive and safety synthesis rewards, \texttt{DriveMind} trains a Soft Actor-Critic (SAC) agent that achieves state-of-the-art performance in CARLA Town 2, with an average speed of 19.4 km/h, 98\% route completion, and near-zero collision speed. It also transfers its semantic objectives zero-shot to real dash-cam data with minimal distributional shift.


\textbf{Future Work.} While \texttt{DriveMind} performs well in simulation and static real-world imagery, several extensions remain. We plan to transition to closed-loop real-world driving, including dynamic rollouts and hardware-in-the-loop testing. Further, we will include automated tuning of reward weights ($\rho_1$, $\rho_2$, $\lambda$), runtime optimization for real-time deployment, and further exploration of safety risks such as segmentation noise, rare-event failures, and reward misalignment.


\clearpage

\section*{Limitations}

While \texttt{DriveMind} advances end-to-end driving with adaptive semantic rewards and kinematic safety, several limitations remain. First, \texttt{DriveMind} relies on accurate bird's-eye-view semantic segmentation and robust VLM embeddings; sensor noise, occlusions, or domain shifts in the segmentation backbone can corrupt both static and dynamic reward signals. Second, its novelty detector uses a fixed $L_2$ distance threshold, which may over-trigger costly prompt generation in cluttered environments or miss slowly evolving hazards. Third, the compact world model provides only one-step predictive foresight, limiting multi-step planning for complex maneuvers. Finally, current evaluations are confined to CARLA simulation and static dash-cam frames due to hardware and integration constraints; closed-loop real-vehicle testing, including hardware-in-the-loop integration and on-road trials, is required to validate performance under real driving dynamics and uncover additional practical failure modes.



\bibliography{ref}  

@article{haarnoja2018soft,
  title={Soft actor-critic algorithms and applications},
  author={Haarnoja, Tuomas and Zhou, Aurick and Hartikainen, Kristian and Tucker, George and Ha, Sehoon and Tan, Jie and Kumar, Vikash and Zhu, Henry and Gupta, Abhishek and Abbeel, Pieter and others},
  journal={arXiv preprint arXiv:1812.05905},
  year={2018}
}

@inproceedings{dosovitskiy2017carla,
  title={{CARLA}: An open urban driving simulator},
  author={Dosovitskiy, Alexey and Ros, German and Codevilla, Felipe and Lopez, Antonio and Koltun, Vladlen},
  booktitle={Conference on robot learning},
  pages={1--16},
  year={2017},
  organization={PMLR}
}

@mastersthesis{markhus2023reinforced,
  title={Reinforced InterFuser for end-to-end autonomous driving in simulated environments},
  author={Markhus, H{\aa}vard Stavn{\aa}s},
  year={2023},
  school={NTNU}
}

@inproceedings{hu2023planning,
  title={Planning-oriented autonomous driving},
  author={Hu, Yihan and Yang, Jiazhi and Chen, Li and Li, Keyu and Sima, Chonghao and Zhu, Xizhou and Chai, Siqi and Du, Senyao and Lin, Tianwei and Wang, Wenhai and others},
  booktitle={Proceedings of the IEEE/CVF conference on computer vision and pattern recognition},
  pages={17853--17862},
  year={2023}
}

@inproceedings{jia2023think,
  title={Think twice before driving: Towards scalable decoders for end-to-end autonomous driving},
  author={Jia, Xiaosong and Wu, Penghao and Chen, Li and Xie, Jiangwei and He, Conghui and Yan, Junchi and Li, Hongyang},
  booktitle={Proceedings of the IEEE/CVF Conference on Computer Vision and Pattern Recognition},
  pages={21983--21994},
  year={2023}
}

@inproceedings{shao2023reasonnet,
  title={Reasonnet: End-to-end driving with temporal and global reasoning},
  author={Shao, Hao and Wang, Letian and Chen, Ruobing and Waslander, Steven L and Li, Hongsheng and Liu, Yu},
  booktitle={Proceedings of the IEEE/CVF conference on computer vision and pattern recognition},
  pages={13723--13733},
  year={2023}
}

@article{wu2022trajectory,
  title={Trajectory-guided control prediction for end-to-end autonomous driving: A simple yet strong baseline},
  author={Wu, Penghao and Jia, Xiaosong and Chen, Li and Yan, Junchi and Li, Hongyang and Qiao, Yu},
  journal={Advances in Neural Information Processing Systems},
  volume={35},
  pages={6119--6132},
  year={2022}
}

@inproceedings{chen2022learning,
  title={Learning from all vehicles},
  author={Chen, Dian and Kr{\"a}henb{\"u}hl, Philipp},
  booktitle={Proceedings of the IEEE/CVF Conference on Computer Vision and Pattern Recognition},
  pages={17222--17231},
  year={2022}
}

@inproceedings{zhang2021end,
  title={End-to-end urban driving by imitating a reinforcement learning coach},
  author={Zhang, Zhejun and Liniger, Alexander and Dai, Dengxin and Yu, Fisher and Van Gool, Luc},
  booktitle={Proceedings of the IEEE/CVF International Conference on Computer Vision},
  pages={15222--15232},
  year={2021}
}

@inproceedings{zhou2023accelerating,
  title={Accelerating reinforcement learning for autonomous driving using task-agnostic and ego-centric motion skills},
  author={Zhou, Tong and Wang, Letian and Chen, Ruobing and Wang, Wenshuo and Liu, Yu},
  booktitle={2023 IEEE/RSJ International Conference on Intelligent Robots and Systems (IROS)},
  pages={11289--11296},
  year={2023},
  organization={IEEE}
}

@article{xu2024drivegpt4,
  title={Drivegpt4: Interpretable end-to-end autonomous driving via large language model},
  author={Xu, Zhenhua and Zhang, Yujia and Xie, Enze and Zhao, Zhen and Guo, Yong and Wong, Kwan-Yee K and Li, Zhenguo and Zhao, Hengshuang},
  journal={IEEE Robotics and Automation Letters},
  year={2024},
  publisher={IEEE}
}

@article{mao2023gpt,
  title={Gpt-driver: Learning to drive with gpt},
  author={Mao, Jiageng and Qian, Yuxi and Ye, Junjie and Zhao, Hang and Wang, Yue},
  journal={arXiv preprint arXiv:2310.01415},
  year={2023}
}

@article{wang2023empowering,
  title={Empowering autonomous driving with large language models: A safety perspective},
  author={Wang, Yixuan and Jiao, Ruochen and Zhan, Sinong Simon and Lang, Chengtian and Huang, Chao and Wang, Zhaoran and Yang, Zhuoran and Zhu, Qi},
  journal={arXiv preprint arXiv:2312.00812},
  year={2023}
}

@inproceedings{chen2024driving,
  title={Driving with llms: Fusing object-level vector modality for explainable autonomous driving},
  author={Chen, Long and Sinavski, Oleg and H{\"u}nermann, Jan and Karnsund, Alice and Willmott, Andrew James and Birch, Danny and Maund, Daniel and Shotton, Jamie},
  booktitle={2024 IEEE International Conference on Robotics and Automation (ICRA)},
  pages={14093--14100},
  year={2024},
  organization={IEEE}
}

@article{tang2024efficient,
  title={Efficient and generalized end-to-end autonomous driving system with latent deep reinforcement learning and demonstrations},
  author={Tang, Zuojin and Chen, Xiaoyu and Li, YongQiang and Chen, Jianyu},
  journal={arXiv preprint arXiv:2401.11792},
  year={2024}
}

@inproceedings{radford2021learning,
  title={Learning transferable visual models from natural language supervision},
  author={Radford, Alec and Kim, Jong Wook and Hallacy, Chris and Ramesh, Aditya and Goh, Gabriel and Agarwal, Sandhini and Sastry, Girish and Askell, Amanda and Mishkin, Pamela and Clark, Jack and others},
  booktitle={International conference on machine learning},
  pages={8748--8763},
  year={2021},
  organization={PmLR}
}

@article{cao2022trustworthy,
  title={Trustworthy safety improvement for autonomous driving using reinforcement learning},
  author={Cao, Zhong and Xu, Shaobing and Jiao, Xinyu and Peng, Huei and Yang, Diange},
  journal={Transportation research part C: emerging technologies},
  volume={138},
  pages={103656},
  year={2022},
  publisher={Elsevier}
}

@article{chen2021interpretable,
  title={Interpretable end-to-end urban autonomous driving with latent deep reinforcement learning},
  author={Chen, Jianyu and Li, Shengbo Eben and Tomizuka, Masayoshi},
  journal={IEEE Transactions on Intelligent Transportation Systems},
  volume={23},
  number={6},
  pages={5068--5078},
  year={2021},
  publisher={IEEE}
}

@article{wang2023efficient,
  title={Efficient reinforcement learning for autonomous driving with parameterized skills and priors},
  author={Wang, Letian and Liu, Jie and Shao, Hao and Wang, Wenshuo and Chen, Ruobing and Liu, Yu and Waslander, Steven L},
  journal={arXiv preprint arXiv:2305.04412},
  year={2023}
}

@inproceedings{zhang2024chatscene,
  title={Chatscene: Knowledge-enabled safety-critical scenario generation for autonomous vehicles},
  author={Zhang, Jiawei and Xu, Chejian and Li, Bo},
  booktitle={Proceedings of the IEEE/CVF Conference on Computer Vision and Pattern Recognition},
  pages={15459--15469},
  year={2024}
}

@article{hazra2024revolve,
  title={REvolve: Reward Evolution with Large Language Models using Human Feedback},
  author={Hazra, Rishi and Sygkounas, Alkis and Persson, Andreas and Loutfi, Amy and Martires, Pedro Zuidberg Dos},
  journal={arXiv preprint arXiv:2406.01309},
  year={2024}
}

@article{baumli2023vision,
  title={Vision-language models as a source of rewards},
  author={Baumli, Kate and Baveja, Satinder and Behbahani, Feryal and Chan, Harris and Comanici, Gheorghe and Flennerhag, Sebastian and Gazeau, Maxime and Holsheimer, Kristian and Horgan, Dan and Laskin, Michael and others},
  journal={arXiv preprint arXiv:2312.09187},
  year={2023}
}

@article{sontakke2023roboclip,
  title={Roboclip: One demonstration is enough to learn robot policies},
  author={Sontakke, Sumedh and Zhang, Jesse and Arnold, S{\'e}b and Pertsch, Karl and B{\i}y{\i}k, Erdem and Sadigh, Dorsa and Finn, Chelsea and Itti, Laurent},
  journal={Advances in Neural Information Processing Systems},
  volume={36},
  pages={55681--55693},
  year={2023}
}

@article{rocamonde2023vision,
  title={Vision-language models are zero-shot reward models for reinforcement learning},
  author={Rocamonde, Juan and Montesinos, Victoriano and Nava, Elvis and Perez, Ethan and Lindner, David},
  journal={arXiv preprint arXiv:2310.12921},
  year={2023}
}

@article{ye2024lord,
  title={Lord: Large models based opposite reward design for autonomous driving},
  author={Ye, Xin and Tao, Feng and Mallik, Abhirup and Yaman, Burhaneddin and Ren, Liu},
  journal={arXiv preprint arXiv:2403.18965},
  year={2024}
}

@article{huang2024vlm,
  title={VLM-RL: A Unified Vision Language Models and Reinforcement Learning Framework for Safe Autonomous Driving},
  author={Huang, Zilin and Sheng, Zihao and Qu, Yansong and You, Junwei and Chen, Sikai},
  journal={arXiv preprint arXiv:2412.15544},
  year={2024}
}

@book{villani2008optimal,
  title={Optimal transport: old and new},
  author={Villani, C{\'e}dric and others},
  volume={338},
  year={2008},
  publisher={Springer}
}

@article{smirnov1939estimation,
  title={On the estimation of the discrepancy between empirical curves of distribution for two independent samples},
  author={Smirnov, Nikolai V},
  journal={Bull. Math. Univ. Moscou},
  volume={2},
  number={2},
  pages={3--14},
  year={1939}
}

@article{deng2023implicit,
  title={Implicit chain of thought reasoning via knowledge distillation},
  author={Deng, Yuntian and Prasad, Kiran and Fernandez, Roland and Smolensky, Paul and Chaudhary, Vishrav and Shieber, Stuart},
  journal={arXiv preprint arXiv:2311.01460},
  year={2023}
}

\clearpage
\begin{center}
    \Large{\bf APPENDICES}
\end{center}
\appendix
\renewcommand{\theequation}{A\arabic{equation}}
\setcounter{equation}{0}
\section{Preliminaries}

\subsection{Partially‐Observable Markov Decision Processes and Soft Actor‑Critic}
We model autonomous driving as a partially‐observable Markov decision process (POMDP) $(\mathcal{S},\mathcal{A},\mathcal{O},\mathcal{T}, R,\gamma)$, where $\mathcal{S}$ is the hidden state space, $\mathcal{A}$ the continuous control actions, and $\mathcal{O}$ the agent’s observations. The goal is to learn a policy $\pi_\phi(a\!\mid\!o)$ that maximizes the expected discounted return
\begin{equation}
  J(\pi_\phi) = \mathbb{E}\Bigl[\sum_{t=0}^{\infty}\gamma^t\,R(s_t,a_t)\Bigr]\,. 
\end{equation}
Soft Actor-Critic (SAC) enhances this by adding an entropy bonus, yielding
\begin{equation}
  J_{\rm SAC}(\pi_\phi)\!=\!\mathbb{E}\Bigl[\sum_{t=0}^\infty\gamma^t\bigl(R(s_t,a_t)+\alpha\mathcal{H}(\pi_\phi(\cdot\mid o_t))\bigr)\Bigr],
\end{equation}
where $\mathcal{H}$ is the policy entropy and $\alpha$ a temperature hyperparameter. SAC alternates between minimizing the soft‐Bellman residual for the Q‐function,
\begin{equation}
  \mathcal{L}_Q = \mathbb{E}_{(o,a,r,o')\sim\mathcal D}
    \bigl(Q_\theta(o,a)-[r + \gamma\,\mathbb{E}_{a'\sim\pi}Q_{\bar\theta}(o',a')]\bigr)^2,    
\end{equation}
and updating the actor by minimizing the KL divergence to the \emph{soft} Q‐values.

\subsection{Contrastive Representation Learning}
Contrastive methods align image and text embeddings in a shared latent space by pulling matched pairs together and pushing mismatches apart.  Given $N$ pairs $\{(x_i,y_i)\}$, encoders $f_I$ and $f_L$ are trained with the InfoNCE loss
\begin{equation}
  \mathcal{L}_{\rm NCE} 
  = -\frac1N\sum_i\log\frac{\exp\!\bigl(\mathrm{sim}(f_I(x_i),f_L(y_i))/\tau\bigr)}
                         {\sum_j\exp\!\bigl(\mathrm{sim}(f_I(x_i),f_L(y_j))/\tau\bigr)},
\end{equation}
where $\mathrm{sim}(u,v)=u^\top v/\|u\|\|v\|$ and $\tau$ is a temperature.  Once frozen, $f_I(x_t)$ and a fixed goal embedding $f_L(y^*)$ yield a dense, zero‐shot reward:
\begin{equation}
r_t^{\rm VLM} = \mathrm{sim}\bigl(f_I(x_t),\,f_L(y^*)\bigr)\in[0,1],    
\end{equation}
quantifying semantic alignment to a natural‐language objective.

\subsection{Chain‑of‑Thought Knowledge Distillation}
Encoder-decoder VLMs generate text from images by first encoding an input $x$ into $z=E_V(x)$, then autoregressively decoding tokens $y=(y_1,\ldots,y_T)$ via
\begin{equation}
  p_\theta(y\mid x)=\prod_{t=1}^T p_\theta(y_t\mid y_{<t},z)\,.    
\end{equation}

To imbue such models with structured, human-like reasoning, we employ chain-of-thought (CoT) knowledge distillation (KD). A large, pretrained teacher model (e.g., GPT-4) first produces both a final target output and an explicit CoT rationale for each training image. The student VLM is then fine-tuned with a composite loss that combines a standard cross-entropy term on the final targets and a KL-divergence term on the teacher’s reasoning distribution:
\[
  \mathcal{L}_{\mathrm{KD}}
    = \mathrm{CE}\bigl(y_{\rm target},\,\hat y\bigr)
      + \lambda\;\mathrm{KL}\!\bigl(p_{\rm student}(r\mid x)\,\big\|\,p_{\rm teacher}(r\mid x)\bigr),
\]
where $r$ denotes the sequence of reasoning tokens. This distillation objective teaches the student not only what to say, but also how to think, yielding a compact model capable of generating on-demand explanations alongside its predictions.

\section{Combined Reward and Training Algorithm}

In this section we present the mathematical formulation of \texttt{DriveMind}’s composite reward, derive each component in detail, and lay out the complete training algorithm with all update rules. We begin by restating the per‐step reward decomposition, then describe the Soft Actor-Critic updates, the world‐model optimization, the chain‐of‐thought distillation for the dynamic VLM, and the novelty detection mechanism. Finally, we provide the complete pseudocode.

\subsection{Reward Decomposition}

\texttt{DriveMind}’s per‐step scalar reward $r_t$ combines four terms:
\begin{equation}
r_t = r_{\mathrm{task},t} \;+\; \rho_1\,R_{\mathrm{synth}}(s_t)
      \;+\;\rho_2\,r^{\mathrm{AICR}}_t
      \;+\;\lambda\,r^{\mathrm{PCFM}}_t,    
\end{equation}
where:
\begin{itemize}
  \item $r_{\mathrm{task},t}$ is the environment‐provided task reward (e.g.\ progress along route).
  \item $R_{\mathrm{synth}}(s_t)$ is the hierarchical safety synthesis term,
  \begin{equation}
      R_{\mathrm{synth}}(s_t)
        = r_{\mathrm{speed}}(s_t)\times f_{\mathrm{center}}(s_t)
          \times f_{\mathrm{angle}}(s_t)\times f_{\mathrm{stability}}(s_t).    
  \end{equation}
  \item $r^{\mathrm{AICR}}_t$ is the Adaptive ideal‐State Contrastive Reward,
  \begin{equation}
      r^{\mathrm{AICR}}_t
        = \alpha\,\cos\bigl(v_t,u^I_t\bigr)
          \;-\;\beta\,\cos\bigl(v_t,u^P_t\bigr),
      \quad \alpha+\beta=1.      
  \end{equation}
  \item $r^{\mathrm{PCFM}}_t$ is the Predictive Contrastive Foresight Module reward,
  \begin{equation}
      r^{\mathrm{PCFM}}_t
        = \cos\bigl(\hat v_{t+1},u^I_t\bigr),
      \quad \hat v_{t+1} = M_\phi(v_t,a_t).      
  \end{equation}
\end{itemize}
The weighting coefficients satisfy $\rho_1\gg \rho_2$ and $\lambda\ll \rho_2$, ensuring that hard safety constraints dominate while semantic guidance and predictive foresight provide nuanced shaping.

\subsection{Soft Actor-Critic Updates}

We train a stochastic policy $\pi_\phi(a|o)$ and a soft Q‐function $Q_\theta(o,a)$ via Soft Actor-Critic. The SAC objective consists of:

\subsubsection{Critic Loss}
\begin{align}
  J_Q(\theta) &= 
    \mathbb{E}_{(o,a,r,o')\sim\mathcal{D}}\Bigl[\tfrac12 \bigl(Q_\theta(o,a) - y\bigr)^2\Bigr], \\
  y &= r + \gamma\,\mathbb{E}_{a'\sim\pi_\phi}\bigl[Q_{\bar\theta}(o',a') - \alpha \log\pi_\phi(a'|o')\bigr].
\end{align}
Here $\bar\theta$ are slowly updated target network parameters via Polyak averaging:
\begin{equation}
  \bar\theta \leftarrow \tau\,\theta + (1-\tau)\,\bar\theta.    
\end{equation}

\subsubsection{Actor Loss}
\begin{equation}
  J_\pi(\phi) 
    = \mathbb{E}_{o\sim\mathcal{D},\,a\sim\pi_\phi}\Bigl[\alpha \log\pi_\phi(a|o) - Q_\theta(o,a)\Bigr].    
\end{equation}
We update $\phi$ by gradient descent and adjust $\alpha$ to match a target entropy $\mathcal{H}_{\mathrm{target}}$.

\subsection{World Model Optimization}

The compact world model $M_\phi\colon\mathbb{R}^k\times\mathcal{A}\to\mathbb{R}^k$ is trained to minimize the one‐step prediction error:
\begin{equation}
  \mathcal{L}_{\mathrm{world}}(\phi)
    = \mathbb{E}_{(v_t,a_t,v_{t+1})\sim\mathcal{D}}
      \bigl\|\;M_\phi(v_t,a_t)\;-\;v_{t+1}\bigr\|_2^2.    
\end{equation}
We perform $U_{\mathrm{WM}}$ world‐model updates for every SAC batch, using Adam with learning rate $\eta_{\mathrm{WM}}$.

\subsection{Chain‐of‐Thought Distillation}

When invoked, the dynamic VLM $\mathrm{VLM}^D$ is fine‐tuned via CoT distillation from a GPT-4 teacher. Given an input image $x$, the teacher produces a target prompt sequence $y_{\rm prompt}$ and an accompanying chain‐of‐thought rationale $y_{\rm CoT}$. We update the student model by minimizing a combined cross‐entropy loss over both the prompt and the reasoning tokens:
\[
\mathcal{L}_{\rm CoT}
  = \mathrm{CE}\bigl(y_{\rm prompt},\,\hat y_{\rm prompt}\bigr)
    + \lambda\,\mathrm{CE}\bigl(y_{\rm CoT},\,\hat y_{\rm CoT}\bigr),
\]
where $\hat y_{\rm prompt}$ and $\hat y_{\rm CoT}$ are the student’s generated prompt and rationale, respectively, and $\lambda$ balances the two terms. This objective ensures the student not only reproduces the correct directives but also the intermediate reasoning steps that led to them.

\subsection{Novelty Detection Formalism}

The bird's-eye-view (BEV) encoder $\mathrm{VLM}^C_I$ maps $\psi(s_t)$ to $v_t\in\mathbb{R}^k$. We maintain a buffer $\mathcal{B}=\{v^{(i)}\}_{i=1}^K$. The novelty score is:
\begin{equation}
\Delta_t = \min_{v'\in\mathcal{B}}\;\|v_t - v'\|_2.    
\end{equation}
If $\Delta_t>\delta$, we invoke the dynamic VLM and update the buffer via FIFO replacement.

\subsection{Complete Training Loop}

\begin{algorithm}[ht]
\caption{\texttt{DriveMind} Training Algorithm}
\begin{algorithmic}[1]
\STATE \textbf{Input:} static VLM $\mathrm{VLM}^C$, dynamic VLM $\mathrm{VLM}^D$, world model $M_\phi$, SAC networks $(\pi_\phi,Q_\theta)$, buffer $\mathcal{D}$, novelty buffer $\mathcal{B}$, hyperparameters $(\rho_1,\rho_2,\lambda,\delta,K)$.
\STATE Initialize $\mathcal{D}, \mathcal{B}$ with random embeddings.
\FOR{environment steps $t=1$ to $T$}
  \STATE Observe $o_t=(\psi(s_t),e_t,w_t)$, compute $v_t=\mathrm{VLM}^C_I(\psi(s_t))$.
  \STATE $\Delta_t\gets \min_{v'\in\mathcal{B}}\|v_t-v'\|_2$.
  \IF{$\Delta_t>\delta$}
    \STATE Generate $(P_t,I_t,\mathrm{CoT}_t)\sim \mathrm{VLM}^D(\psi(s_t))$.
    \STATE Compute $u^P_t,u^I_t$ via text encoder, cache in $\mathcal{B}$.
    \STATE Optionally perform $\mathcal{L}_{\mathrm{KD}}$ update on $\mathrm{VLM}^D$.
  \ELSE
    \STATE Reuse $u^P_t,u^I_t$ from last trigger.
  \ENDIF
  \STATE Compute $R_{\mathrm{synth}}(s_t)$, $r^{\mathrm{AICR}}_t$, $\hat v_{t+1}=M_\phi(v_t,a_{t})$, $r^{\mathrm{PCFM}}_t$.
  \STATE Sample action $a_t\sim\pi_\phi(a|o_t)$, apply, observe $r_{\mathrm{task},t},s_{t+1}$.
  \STATE Compute total reward $r_t$ and store $(o_t,a_t,r_t,o_{t+1})$ in $\mathcal{D}$.
  \IF{time to update SAC}
    \STATE Sample mini-batch from $\mathcal{D}$.
    \STATE Update $Q_\theta$ by minimizing $J_Q(\theta)$.
    \STATE Update $\pi_\phi$ by minimizing $J_\pi(\phi)$.
    \STATE Update world model $M_\phi$ via $\mathcal{L}_{\mathrm{world}}$.
    \STATE Update target networks $\bar\theta\leftarrow\tau\theta+(1-\tau)\bar\theta$.
  \ENDIF
\ENDFOR
\end{algorithmic}
\end{algorithm}


\section{Experiments and Results}
\label{sec:result}

\subsection{Experimental Setup}
\subsubsection{Reinforcement Learning Configuration}

In our experiments, the \texttt{DriveMind} agent is trained with Soft Actor-Critic (SAC) as implemented in Stable‑Baselines3.  At each timestep $t$, the agent receives an observation
\begin{equation}
o_t = \bigl(\psi(s_t),\; e_t,\; w_t\bigr),    
\end{equation}
where:
\begin{itemize}
  \item $\psi(s_t)\in\mathbb{R}^{H\times W\times C}$ is the bird’s‑eye‑view (BEV) semantic segmentation image of the environment.
  \item $e_t = \bigl(v_t,\;\theta_t,\;\omega_t\bigr)\in\mathbb{R}^3$ is the ego‑state vector, comprising the vehicle’s longitudinal speed $v_t$, steering angle $\theta_t$, and yaw rate $\omega_t$.
  \item $w_t = (x_{t+1},y_{t+1},\,\dots\,,x_{t+K},y_{t+K})\in\mathbb{R}^{2K}$ are the coordinates of the next $K$ waypoints along the planned route, in the vehicle’s local frame.
\end{itemize}

\paragraph{Action Space}  
The continuous action $a_t\in\mathcal{A}=[-1,1]^2$ controls steering and combined throttle/brake.  We decode:
\begin{align}
  \delta_t &= a_t^{(1)} \cdot \delta_{\max},\\
  \tau_t    &= \max\bigl(a_t^{(2)}, 0\bigr),\quad
  b_t       = -\min\bigl(a_t^{(2)}, 0\bigr),
\end{align}
where $\delta_t$ is the steering angle (scaled by maximum $\delta_{\max}$), $\tau_t$ the throttle in $[0,1]$, and $b_t$ the brake in $[0,1]$.

\paragraph{Termination Conditions.}  
An episode terminates if any of the following occurs:
\begin{enumerate}
  \item Collision with static obstacles, other vehicles, or pedestrians.
  \item Lateral deviation from the lane centerline exceeds $d_{\max}$ (e.g.\ $3\text{\,m}$).
  \item Vehicle speed remains below $v_{\min}$ (e.g.\ $1\text{\,km/h}$) for more than $T_{\rm stuck}$ (e.g.\ $90$s).
\end{enumerate}

\paragraph{Network Architectures.}  
Both the actor $\pi_\phi(a\mid o)$ and critic $Q_\theta(o,a)$ share a common feature extractor that fuses visual and proprioceptive inputs:
\begin{equation}
z_t = \bigl[f_{\rm bev}\bigl(\psi(s_t)\bigr)\;\|\;f_{\rm ego}(e_t)\;\|\;f_{\rm wp}(w_t)\bigr]\;\in\;\mathbb{R}^D.    
\end{equation}
\begin{itemize}
  \item \textbf{BEV encoder $f_{\rm bev}\colon\mathbb{R}^{H\times W\times C}\to\mathbb{R}^{d_v}$.}  A 6‑layer convolutional network with ReLU activations:
  \begin{equation}
      \begin{aligned}
        &\text{Conv2D}(C,32,8,4)\to\mathrm{ReLU},\quad \text{Conv2D}(32,64,4,2)\to\mathrm{ReLU},\\
        &\text{Conv2D}(64,64,3,1)\to\mathrm{ReLU},\quad \text{Flatten}\,\to\;\text{FC}(1024)\to\mathrm{ReLU}\;\to\;\mathbb{R}^{d_v}.
      \end{aligned}      
  \end{equation}
  \textbf{Ego‐state encoder $f_{\rm ego}$.}  A two‐layer MLP that takes a 3‐dimensional ego state vector $(\text{steering},\;\text{throttle},\;\text{speed})$ as input and outputs a $d_e$‐dimensional embedding: 
  \begin{equation}
f_{\rm ego}\colon \mathbb{R}^3\to\mathbb{R}^{d_e}.      
  \end{equation}

  \textbf{Waypoint encoder $f_{\rm wp}$.}  
A two‐layer MLP mapping $\mathbb{R}^{2K}$ (the next $K$ $(x,y)$ waypoints) to $\mathbb{R}^{d_w}$: FC(128)$\to$ReLU$\to$FC($d_w$)$\to$ReLU.

\end{itemize}
The concatenated feature $z_t\in\mathbb{R}^{d_v + d_e + d_w}$ is fed into:
\begin{itemize}
  \item \textbf{Critic network $Q_\theta$}: a three‑layer MLP mapping $\bigl(z_t, a_t\bigr)\to\mathbb{R}$.
  \item \textbf{Actor network $\pi_\phi$}: a three‑layer MLP mapping $z_t$ to the parameters of a diagonal Gaussian over $\mathcal{A}$.
\end{itemize}

\paragraph{SAC Objective.}  
We optimize the soft Q‑function by minimizing the Bellman residual:
\begin{equation}
J_Q(\theta)
= \mathbb{E}_{(o,a,r,o')\sim\mathcal D}\Bigl[\tfrac12\bigl(Q_\theta(o,a) - y(r,o')\bigr)^2\Bigr],    
\end{equation}
with target:
\begin{equation}
y(r,o') = r + \gamma\,\mathbb{E}_{a'\sim\pi_\phi}\bigl[\,Q_{\bar\theta}(o',a') - \alpha\log\pi_\phi(a'\mid o')\bigr]
\end{equation}
and update the policy by minimizing
\begin{equation}
J_\pi(\phi)
= \mathbb{E}_{o\sim\mathcal D}\Bigl[\mathbb{E}_{a\sim\pi_\phi}\bigl[\alpha\log\pi_\phi(a\mid o) - Q_\theta(o,a)\bigr]\Bigr].    
\end{equation}
All networks are trained off‑policy using a replay buffer $\mathcal D$, with target networks $Q_{\bar\theta}$ periodically updated.  

\subsubsection{Vision Language Model Configurations}

\texttt{DriveMind} relies on two distinct VLMs: a frozen, high‑capacity CLIP model for static semantic anchoring and a compact encoder-decoder model for on‑demand prompt generation.

\paragraph{Static Contrastive VLM.}  
For stable, per‑step embedding of bird’s‑eye‑view observations, we use the OpenCLIP ViT‑bigG‑14 model.  Its image encoder partitions each input frame (resized to $224\times224$px) into non‑overlapping $14\times14$px patches, projects each patch into a 1\,024‑dimensional token, and processes the resulting sequence through a 24‑layer Vision Transformer.  The final CLS token yields a $k=2{,}048$‑dimensional embedding
\begin{equation}
v_t \;=\;\mathrm{VLM}^C_I\bigl(\psi(s_t)\bigr)\;\in\;\mathbb{R}^{2048}.    
\end{equation}
Similarly, two fixed text prompts $l^P,l^I$ are encoded once into
\begin{equation}
u^P=\mathrm{VLM}^C_L(l^P),\quad u^I=\mathrm{VLM}^C_L(l^I),    
\end{equation}
each $\in\mathbb{R}^{2048}$.  All weights of ViT‑bigG‑14 (image and text branches) remain frozen during \texttt{DriveMind} training and evaluation, ensuring a consistent semantic reference frame.

\paragraph{Dynamic Encoder-Decoder VLM.}  
To generate context‑specific “present” and “ideal” prompts only when needed, we employ SmolVLM‑256M, a lightweight transformer with approximately 256 million parameters.  SmolVLM’s vision encoder mirrors a ViT‑Tiny architecture (patch size $16\times16$, input resolution $128\times128$ px), producing a $d=512$‑dimensional latent $z_t$.  Its autoregressive text decoder is a 6‑layer Transformer that, when triggered, generates:
\begin{equation}
P_t,\;I_t,\;\mathrm{CoT}_t
\;\sim\;p_\theta\bigl(\cdot\mid z_t\bigr),    
\end{equation}
where $P_t$ and $I_t$ are the natural‑language “present” and “ideal” prompts, and $\mathrm{CoT}_t$ is the chain‑of‑thought rationale.  These are re‑embedded via the SmolVLM text encoder into
\begin{equation}
u^P_t,\;u^I_t \;\in\;\mathbb{R}^{512},    
\end{equation}
and cached until the novelty detector (Eq.~\ref{eq:novelty_score}) signals a scene change.  SmolVLM is pretrained via chain‑of‑thought knowledge distillation from a GPT-4 teacher, minimizing a combined cross‑entropy and KL loss that aligns both the prompt outputs and their intermediate rationales.

\paragraph{Text‑Prompt Cache Behavior.}  
A fixed‐size cache $\{(u^P_{t_i},u^I_{t_i})\}_{i=1}^K$ stores the last $K$ dynamic embeddings.  Upon each SmolVLM invocation, the oldest entry is evicted and replaced, ensuring that subsequent timesteps reuse these context‑specific prompts without redundant decoding.  This design keeps real‑time inference costs low while retaining up‑to‑date, scene‑adaptive semantic objectives.

\subsection{Evaluation Metrics}

We evaluate \texttt{DriveMind} along five key axes (on CARLA), each summarized by a scalar metric:

\begin{itemize}
  \item \textbf{Average Speed (AS)}: the time‑averaged vehicle speed over an episode of length $T$,
  \begin{equation}
    \mathrm{AS} \;=\;\frac{1}{T}\sum_{t=1}^{T} v_t
    \;=\;\frac{D_{\mathrm{tot}}}{T},      
  \end{equation}
  where $v_t$ is the speed at timestep $t$ and 
  $D_{\mathrm{tot}} = \sum_{t=1}^{T}\|p_t - p_{t-1}\|_2$
  is the total distance traveled.

  \item \textbf{Route Completion Rate (RC)}: the fraction of episodes in which the agent reaches its goal before termination,
  \begin{equation}
    \mathrm{RC}
      = \frac{1}{N_{\rm ep}}
        \sum_{i=1}^{N_{\rm ep}}
        \mathbf{1}\{\text{episode }i\text{ reaches its goal}\}\,\in[0,1],      
  \end{equation}
  where $N_{\rm ep}$ is the total number of episodes.

  \item \textbf{Total Distance (TD)}: the cumulative path length per episode, averaged over all episodes,
  \begin{equation}
    \mathrm{TD}
      = \frac{1}{N_{\rm ep}}
        \sum_{i=1}^{N_{\rm ep}}
        \Bigl(\sum_{t=1}^{T_i}\|p_t^{(i)} - p_{t-1}^{(i)}\|_2\Bigr)\,,      
  \end{equation}
  with $T_i$ the length of episode $i$.

  \item \textbf{Collision Speed (CS)}: the mean speed at collision events,
  \begin{equation}
    \mathrm{CS}
      = \frac{1}{N_{\rm col}}
        \sum_{j=1}^{N_{\rm col}}
        v_{\rm col}^{(j)},      
  \end{equation}
  where $v_{\rm col}^{(j)}$ is the vehicle’s speed at the $j$‑th collision and $N_{\rm col}$ the total number of collisions.

  \item \textbf{Success Rate (SR)}: the fraction of episodes that both reach the goal and incur no collisions,
  \begin{equation}
    \mathrm{SR}
      = \frac{1}{N_{\rm ep}}
        \sum_{i=1}^{N_{\rm ep}}
        \mathbf{1}\{\text{episode }i\text{ succeeds without collision}\}
      \;\in[0,1].      
  \end{equation}
\end{itemize}

\subsection{Baselines}

To provide a comprehensive evaluation of \texttt{DriveMind}, we compare against fourteen state‑of‑the‑art reward methods, organized into three categories.  For each, we implement both SAC and PPO where applicable, holding the underlying network architectures and training hyperparameters identical to \texttt{DriveMind}.

\subsubsection{Expert‑Designed Reward Methods}

\paragraph{TIRL‑SAC.}  
TIRL‑SAC \cite{cao2022trustworthy} represents the simplest possible reward: a binary penalty for collisions.  At each step~$t$, the agent receives 
\begin{equation}
R_{\rm TIRL}(s_t,a_t) =
\begin{cases}
-1, & \text{if a collision occurs at time }t,\\
0,  & \text{otherwise.}
\end{cases}    
\end{equation}
This design provides no positive reinforcement for progress, forcing the agent to discover safe behaviors purely by avoiding negative signals.  In practice, this extreme sparsity leads to very slow or degenerate learning: without any incentive to move forward, the optimal strategy is often to stay stationary.  We implement TIRL‑SAC using the standard SAC update rules, with the critic trained to minimize the soft Bellman residual against the sparse $R_{\rm TIRL}$, and the actor updated to maximize a combination of Q‑values and policy entropy.  All other hyperparameters (learning rate, replay buffer size, batch size, target update rate) match those used for \texttt{DriveMind}.

\paragraph{TIRL‑PPO.}  
TIRL‑PPO \cite{cao2022trustworthy} substitutes the PPO algorithm in place of SAC, but retains the identical binary reward:
\begin{equation}
R_{\rm TIRL\text{-}PPO}(s_t,a_t) =
\begin{cases}
-1, & \text{if collision at }t,\\
0,  & \text{otherwise.}
\end{cases}    
\end{equation}
PPO optimizes
\begin{equation}
J_{\rm PPO}(\phi) = \mathbb{E}\Bigl[\min\bigl(r_t(\phi)\,\hat A_t,\ \mathrm{clip}(r_t(\phi),1-\epsilon,1+\epsilon)\,\hat A_t\bigr)\Bigr],    
\end{equation}
where $r_t(\phi)=\pi_\phi(a_t|s_t)/\pi_{\phi_{\rm old}}(a_t|s_t)$ and $\hat A_t$ is the estimated advantage using $R_{\rm TIRL\text{-}PPO}$.  As with TIRL‑SAC, the absence of any positive driving reward causes PPO to converge to near‐zero speed policies, highlighting the limitations of purely negative feedback.

\paragraph{Chen‑SAC.}  
Chen‑SAC \cite{chen2021interpretable} employs a richer, hand‑designed reward that balances collision avoidance with smooth, forward driving and lane‐keeping:
\begin{equation}
R_{\rm Chen}(s_t,a_t)
= -\,w_{\rm coll}\,\mathbf{1}_{\rm coll}(s_t)
  + w_{\rm speed}\,\frac{v_t}{v_{\max}}
  - w_{\rm lane}\,\Delta_{\rm lane}(s_t)
  - w_{\rm steer}\,\lvert\omega_t\rvert.    
\end{equation}
Here $v_t$ is the vehicle speed, $\Delta_{\rm lane}$ the lateral offset from centerline, and $\omega_t$ the steering angle.  Typical weights are set to $(w_{\rm coll}=10,\ w_{\rm speed}=1,\ w_{\rm lane}=2,\ w_{\rm steer}=0.5)$.  The collision indicator $\mathbf{1}_{\rm coll}$ carries the heaviest penalty to enforce safety.  The SAC critic and actor updates proceed as usual, but driven by this composite signal: positive reward for forward motion, negative for deviations and steering effort.  In practice, Chen‑SAC learns nonzero speeds yet often oscillates in lane‐centering, illustrating the difficulty of balancing multiple hand‑tuned terms.

\paragraph{ASAP‑RL‑PPO.}  
ASAP‑RL‑PPO \cite{wang2023efficient} focuses on progress and overtaking incentives under PPO:
\begin{equation}
R_{\rm ASAP}(s_t,a_t)
= \gamma_{\rm prog}\,\Delta d_t
  + \gamma_{\rm overtake}\,\mathbf{1}_{\rm overtake}(s_t)
  - \gamma_{\rm coll}\,\mathbf{1}_{\rm coll}(s_t)
  - \gamma_{\rm offroad}\,\mathbf{1}_{\rm offroad}(s_t).    
\end{equation}
Here $\Delta d_t$ is the distance advanced along the planned route, $\mathbf{1}_{\rm overtake}$ flags successful passes of slower traffic, and the two penalty indicators discourage collisions and leaving the drivable area.  Weight coefficients are chosen as $(\gamma_{\rm prog}=1,\ \gamma_{\rm overtake}=2,\ \gamma_{\rm coll}=5,\ \gamma_{\rm offroad}=3)$.  PPO’s clipped surrogate objective encourages stable policy improvements based on these incentives.  ASAP‑RL‑PPO tends to favor higher speeds and overtakes but can be overly aggressive, leading to elevated collision rates.

\paragraph{ChatScene‑SAC.}  
ChatScene‑SAC \cite{zhang2024chatscene} constructs a smoothness‐focused reward:
\begin{equation}
R_{\rm CS}(s_t,a_t)
= -\,\lambda_1\bigl|\ddot x_t\bigr|
  -\,\lambda_2\lvert a_{y,t}\rvert
  -\,\lambda_3\,\mathbf{1}_{\rm coll}(s_t)
  + \epsilon,    
\end{equation}
where $\ddot x_t$ is longitudinal acceleration, $a_{y,t}$ lateral acceleration, and $\epsilon=0.1$ provides a constant learning signal.  Typical weights $(\lambda_1,\lambda_2,\lambda_3)=(0.5,0.5,5)$ penalize abrupt motions and collisions.  Using SAC, this reward yields comfortable but often overly conservative driving, as large safety penalties dominate.

\paragraph{ChatScene‑PPO.}  
The same smoothness‐oriented reward $R_{\rm CS}$ is paired with PPO.  PPO’s trajectory batching and clipping help stabilize training but do not fully overcome the conservative bias inherent in the reward design.

\subsubsection{LLM‑Designed Reward Methods}

\paragraph{Revolve.}  
Revolve \cite{hazra2024revolve} leverages a Large Language Model to \emph{generate} candidate reward functions in Python.  The process alternates between prompting the LLM for new code snippets $f_\theta$, evaluating them on a small set of human‐annotated driving trajectories, and selecting the highest‐scoring functions via an evolutionary algorithm.  The resulting reward $R_{\rm Revolve}(s_t,a_t)=f_{\theta^*}(s_t,a_t)$ can include complex conditionals, time‐based terms, or logical combinations of sensors.  When used with SAC or PPO, Revolve often produces unconventional but effective reward structures; however, its reliance on automated code search can introduce brittleness if the LLM generates invalid or overfitted snippets.

\paragraph{Revolve‑auto.}  
Revolve‑auto \cite{hazra2024revolve} extends Revolve by \emph{automatically} labeling its own trajectories for subsequent reward refinement.  After the initial code is selected, the LLM is re‑prompted with performance summaries and asked to generate \emph{refinements} to $f_\theta$.  This loop continues for several rounds, yielding $f_{\theta'}$ that adapts to the agent’s evolving behavior.  While this self‑training can improve alignment with driving objectives, it risks reinforcing unintended biases if the feedback loop is not carefully constrained.

\subsubsection{VLM‑Designed Reward Methods}

\paragraph{VLM‑SR.}  
VLM‑SR \cite{baumli2023vision} uses the CLIP model to compute cosine similarity between the current image embedding $f_I(x_t)$ and a fixed “goal” text embedding $f_L(y^*)$.  Formally,
\begin{equation}
R_{\rm VLM\text{-}SR}(s_t)
= \frac{\exp\!\bigl(\mathrm{sim}(f_I(x_t),f_L(y^*))/\tau\bigr)}
       {\sum_{j}\exp\!\bigl(\mathrm{sim}(f_I(x_t),f_L(y_j))/\tau\bigr)},    
\end{equation}
where $\{y_j\}$ includes negative distractor prompts and $\tau$ is a temperature.  This yields a dense but purely semantic reward, encouraging the agent to produce images “close” to the language goal.  In practice, VLM‑SR alone struggles in continuous driving, as semantic alignment does not directly capture kinematic feasibility.

\paragraph{RoboCLIP.}  
RoboCLIP \cite{sontakke2023roboclip} extends VLM‑SR to video by averaging per‐frame similarities over each episode and penalizing deviations from a task descriptor:
\begin{equation}
R_{\rm RoboCLIP}(s_{0:T})
= \frac{1}{T}\sum_{t=0}^{T-1}
  -\bigl(\mathrm{sim}(f_I(x_t),f_L(y^*)) - 1\bigr)^2.    
\end{equation}
This end‐of‐episode signal can guide off‐policy RL but provides no intermediate shaping, leading to slow convergence in driving tasks.

\paragraph{VLM‑RM.}  
VLM‑RM \cite{rocamonde2023vision} formulates reward as the projection of the current embedding $v_t$ onto the vector from a baseline text $u_{\rm base}$ to a target text $u_{\rm target}$:
\begin{equation}
R_{\rm RM}(s_t)
= (v_t - u_{\rm base})^\top (u_{\rm target} - u_{\rm base}).    
\end{equation}
By measuring progress along the semantic direction, VLM‑RM yields a continuous reward.  However, without negative goals, it may reward ambiguous or unsafe states that partially match the target description.

\paragraph{LORD.}  
LORD \cite{ye2024lord} focuses exclusively on negative-language prompts (e.g.\ “a collision is occurring”).  Its reward is
\begin{equation}
R_{\rm LORD}(s_t)
= -\,\mathrm{sim}\bigl(f_I(x_t),f_L(y_{\rm neg})\bigr),    
\end{equation}
penalizing any semantic resemblance to unsafe conditions. While effective at discouraging collisions, LORD alone provides no positive guidance toward desired behaviors.

\paragraph{LORD‑Speed.}  
LORD‑Speed \cite{ye2024lord} augments LORD with a continuous speed incentive:
\begin{equation}
R_{\rm LORD\text{-}Sp}(s_t)
= -\,\mathrm{sim}\bigl(f_I(x_t),f_L(y_{\rm neg})\bigr)
  + \kappa\,\frac{v_t}{v_{\max}},    
\end{equation}
where $\kappa>0$ trades off semantic safety penalties against forward progress.  This hybrid can drive non-trivial motion but often at the cost of occasional unsafe accelerations.

\paragraph{VLM‑RL.}  
VLM‑RL \cite{huang2024vlm} combines the contrasting language‑goal paradigm (positive and negative prompts) with SAC/RL to produce semantic rewards.  VLM‑RL freezes its VLM, encodes both “present” and “ideal” prompts, and computes
\begin{equation}
R_{\rm VLM\text{-}RL}(s_t)
= \alpha\,\cos(v_t,u^I) - \beta\,\cos(v_t,u^P),    
\end{equation}
without any hierarchical or predictive extensions.

\subsubsection{Driving Environments and Scenarios}

We evaluate \texttt{DriveMind} in two complementary domains: (i) simulated towns in CARLA for controlled, repeatable benchmarks, and (ii) real‐world street scenes drawn from the BDD100K dataset to assess domain transfer.

\paragraph{Simulated CARLA Towns.}  
Let $\mathcal{T} = \{\mathrm{Town}_1,\dots,\mathrm{Town}_5\}$ denote the set of CARLA maps used.  During training, the agent is exposed exclusively to $\mathrm{Town}_2$, characterized by a compact European‐style layout of residential blocks, commercial zones, single‐lane roads, and signalized intersections.  To formally define each episode’s environment, we sample a \emph{scenario} by:
\begin{equation}
s = \bigl(T,\,D,\,R\bigr)\in\mathcal{T}\times\mathcal{D}\times\mathcal{R},    
\end{equation}
where:
\begin{itemize}
  \item $T\in\mathcal{T}$ is the underlying town map.  
  \item $D\in\mathcal{D}=\{\text{empty},\,\text{regular},\,\text{dense}\}$ is the traffic density level.  In “empty” scenarios no other vehicles are spawned.  In “regular,” 20 vehicles operate under CARLA’s traffic manager.  In “dense,” 40 vehicles are deployed.
  \item $R\in\mathcal{R}$ is a navigation route, obtained by selecting two distinct spawn points $p_{\rm start},p_{\rm goal}$ from the set of predefined drivable waypoints $\mathcal{P}$ and computing the shortest path via $A^{*}$:
  \begin{equation}
      R = \mathrm{A}^{*}\bigl(p_{\rm start},\,p_{\rm goal}\bigr).      
  \end{equation}
\end{itemize}
Each episode continues until the vehicle has traversed a total distance of $L_{\max}=3000$m or a termination condition (collision, lane departure, or stuck) is met.  To assess generalization, we evaluate the trained policy on $T\in\{\mathrm{Town}_1,\mathrm{Town}_3,\mathrm{Town}_4,\mathrm{Town}_5\}$ under all three density levels $D$.

\paragraph{Real‐World BDD100K Scenes.}  
To quantify domain transfer, we collect a set of $N=10{,}000$ bird’s‐eye‐view (BEV) projections from the BDD100K validation split.  Let  
\begin{equation}
\mathcal{I} = \{\,I_i \in \mathbb{R}^{H\times W\times 3}\mid i=1,\dots,N\}    
\end{equation}
be the set of images, each paired with ground‐truth lane and vehicle masks.  We define an evaluation scenario for each $I_i$ by treating it as a static scene with $D=\text{empty}$ and no navigation route.  The policy executes a single control step $a_i$ from its observation $o_i=(I_i,e_i,\varnothing)$, and we record the positive and negative VLM similarity scores. 
\begin{equation}
p_i = \cos\bigl(f_I(I_i),\,u^I\bigr),\quad n_i=\cos\bigl(f_I(I_i),\,u^P\bigr). 
\end{equation}
To measure distributional shift, we compare $\{p_i\},\{n_i\}$ against their CARLA Town2 counterparts $\{p_t\},\{n_t\}$ via the Wasserstein distance $W$ and the Kolmogorov-Smirnov statistic $D_{\mathrm{KS}}$:
\begin{equation}
W\bigl(p_i,p_t\bigr),\;D_{\mathrm{KS}}\bigl(p_i,p_t\bigr),\quad
W\bigl(n_i,n_t\bigr),\;D_{\mathrm{KS}}\bigl(n_i,n_t\bigr).    
\end{equation}

\paragraph{Scenario Sampling and Metrics.}  
During both simulated and real‐world evaluation, scenarios are drawn uniformly from their respective sets $\mathcal{T}\times\mathcal{D}\times\mathcal{R}$ or $\mathcal{I}$.  Performance is reported as mean $\pm$ standard deviation over three random seeds (simulated) or three disjoint image subsets (real).  In CARLA, we compute driving efficiency metrics (average speed, route completion, distance traveled) and safety metrics (collision rate, collision speed), while in BDD100K we report the mean positive/negative score and the distributional distances $(W,D_{\mathrm{KS}})$ to quantify domain gap.

\subsubsection{Navigation and Novelty Triggering}

\texttt{DriveMind} generates navigation goals by sampling random routes over a predefined set of spawn points, and invokes the lightweight SmolVLM only when the current scene embedding deviates significantly from past experience.

Consider the finite set of drivable spawn points
\begin{equation}
\mathcal{P} = \{p_1,\ldots,p_M\}\subset\mathbb{R}^2,    
\end{equation}
where each corresponds to a permissible start or goal location in the CARLA map.  At the beginning of each episode (or immediately upon completing a route), we draw two distinct points by:
\begin{equation}
p_{\mathrm{start}},\,p_{\mathrm{goal}}\;\overset{\mathrm{i.i.d.}}{\sim}\;\mathrm{Uniform}(\mathcal{P}),\quad p_{\mathrm{start}}\neq p_{\mathrm{goal}}.  
\end{equation}
We then compute a shortest path $R$ between $p_{\mathrm{start}}$ and $p_{\mathrm{goal}}$ using the A* algorithm over the navigable graph of waypoints:
\begin{equation}
  R \;=\;\mathsf{A}^*\bigl(p_{\mathrm{start}},\,p_{\mathrm{goal}}\bigr), 
  \quad 
  L(R) = \sum_{i=1}^{|R|-1} \bigl\|R_i - R_{i+1}\bigr\|_2,
  \label{eq:route_length}
\end{equation}
where $L(R)$ is the Euclidean length of the polyline $R=(R_1,\dots,R_{|R|})$.  If $L(R)$ exceeds a pre‐specified maximum $L_{\max}$ (e.g.\ 3 km), we reject and resample $(p_{\mathrm{start}},p_{\mathrm{goal}})$ until $L(R)\le L_{\max}$.  By continuously re‐sampling upon completion or termination, the agent experiences diverse navigation challenges within a bounded episode length.

To avoid the computational overhead of running the VLM encoder-decoder (SmolVLM) at every time step, \texttt{DriveMind} employs a novelty detector on the static VLM embeddings.  Let the static model’s vision encoder be:
\begin{equation}
v_t = \mathrm{VLM}^C_I\bigl(\psi(s_t)\bigr)\in\mathbb{R}^k,    
\end{equation}
and maintain a cache 
$\mathcal{B}=\{v^{(1)},\ldots,v^{(K)}\}$ 
of the $K$ most recent embeddings.  We define the novelty score
\begin{equation}
  \Delta_t 
    = \min_{v'\in\mathcal{B}} \;\bigl\lVert v_t - v'\bigr\rVert_2.
  \label{eq:novelty_score}
\end{equation}
A fixed threshold $\delta>0$ determines whether the scene is “novel”:
\begin{equation}
\text{if }\Delta_t > \delta,\quad\text{then invoke SmolVLM on }\psi(s_t)\text{ to generate new prompts;}    
\end{equation}
otherwise, reuse the last cached dynamic embeddings $(u^P_t,u^I_t)$.  Upon invocation, we update the cache via a sliding‐window policy:
\begin{equation}
\mathcal{B} \;\leftarrow\; \bigl(\mathcal{B}\setminus\{v^{(1)}\}\bigr)\;\cup\;\{v_t\},    
\end{equation}
dropping the oldest embedding and appending the current one.  In practice, $\delta$ is chosen as a high quantile of the distribution of $\Delta_t$ observed during a warm‐up phase, ensuring that SmolVLM is called only for genuinely unfamiliar or rare scenes.  This mechanism keeps the per‐step complexity low while granting \texttt{DriveMind} the ability to adapt its semantic objectives precisely when the driving context changes.

\subsection{Implementation Details}

All components of \texttt{DriveMind} are implemented in PyTorch and seamlessly integrated with Stable‐Baselines3 to leverage its robust Soft Actor-Critic (SAC) implementation, efficient replay buffers, and distributed training utilities. Our project is organized as a modular codebase with clearly separated packages for perception, semantic reasoning, world modeling, reinforcement learning, and utilities. The \texttt{perception} module contains the BEV projection pipeline: raw camera images and LiDAR point clouds are first preprocessed using a custom C++/CUDA extension to accelerate the top‐down projection, after which a lightweight semantic segmentation network (based on DeepLabV3) produces multi‐channel BEV masks for lanes, vehicles, pedestrians, and static obstacles. These BEV masks are normalized and resized to $224\times224$ pixels and passed through the frozen CLIP ViT‐bigG‐14 image encoder to produce the static embedding $v_t$. The \texttt{semantic\_reasoning} package wraps both the frozen contrastive static VLM and the dynamic encoder-decoder VLM (SmolVLM); it manages the novelty buffer $\mathcal{B}$, computes the $L_2$ novelty score $\Delta_t$, and handles asynchronous prompt generation via a background thread to avoid blocking the main control loop. Chain‐of‐thought distillation logic resides in a dedicated trainer class that performs an online per-sample distillation using output generated from GPT‐4o-mini API, also handling text tokenization, teacher, student loss computation, and periodic checkpointing. The \texttt{world\_model} package implements the compact MLP predictor $M_\phi$, complete with custom loss wrappers to compute prediction error and integrate it with the RL replay buffer for joint sampling. Finally, the \texttt{rl\_agent} module configures the SAC actor and critic networks, each comprising a shared feature extractor that fuses BEV embeddings with ego‐state and waypoint embeddings, alongside target network management, entropy tuning, and optimizer hooks. All modules share a common configuration system based on YAML files, allowing reproducible experiments via versioned config snapshots. Logging and visualization employ TensorBoard and Weights \& Biases, capturing per‐step metrics, network gradients, and video rollouts for qualitative inspection.

\paragraph{Reinforcement Learning Setup and Hyperparameters}
\texttt{DriveMind}’s reinforcement learning backbone uses Soft Actor-Critic (SAC) following the principles of Stable-Baselines3, extended with our custom composite reward integration. SAC is configured with separate Adam optimizers for the actor and critic networks, each using a learning rate of $3\times10^{-4}$. The critic’s Q-function approximators minimize the squared Bellman residual, and target networks are updated using Polyak averaging with $\tau=5\times10^{-3}$. The discount factor $\gamma$ is set to 0.99 to balance long-term planning and stable learning. Entropy regularization is used to encourage exploration, following standard SAC behavior, although automatic entropy tuning is not explicitly enabled. Each training step samples a mini-batch of 64 transitions from a replay buffer of capacity $5\times10^{4}$, performing one gradient update per environment interaction. We additionally set the composite reward weights as \(\rho_1=1.0\) (hierarchical safety fusion), \(\rho_2=1.0\) (adaptive contrastive term), and \(\lambda=0.05\) (predictive foresight).

The agent’s architecture shares a three-branch feature extractor: a 6-layer CNN processes bird's-eye-view (BEV) images, a two-layer MLP processes the 3-dimensional ego-state vector, and another two-layer MLP handles the 20-dimensional waypoint input. These feature embeddings are concatenated and passed to separate three-layer MLPs for the actor and critic heads. Training uses mixed-precision (PyTorch AMP) for efficiency. Model checkpoints are evaluated every $5\times10^{3}$ environment steps over fixed routes, aggregating metrics including average speed, route completion, distance traveled, collision speed, and success rate. All experiments fix random seeds for Python, NumPy, PyTorch, and the simulator environment to ensure reproducibility.

\begin{table}[h]
    \centering
    \caption{\centering Summary of Key RL Hyperparameters Used in \texttt{DriveMind} Experiments}
    \begin{tabular}{ll}
    \toprule
    \textbf{Hyperparameter} & \textbf{Value} \\
    \midrule
    Learning rate & $3\times10^{-4}$ \\
    Discount factor $\gamma$ & $0.99$ \\
    Target update coefficient $\tau$ & $5\times10^{-3}$ \\
    Replay buffer size & $5\times10^{4}$ \\
    Batch size & $64$ \\
    Policy noise & $0.2$ \\
    Noise clip & $0.5$ \\
    Actor update frequency & $2$ steps \\
    Critic target update frequency & $2$ steps \\
    Exploration noise standard deviation & $0.1$ \\
    Policy update frequency & Every $2$ steps \\
    Evaluation frequency & Every $5\times10^{3}$ steps \\
    \bottomrule
    \end{tabular}
    \vspace{2px}
\end{table}

\begin{figure}[ht]
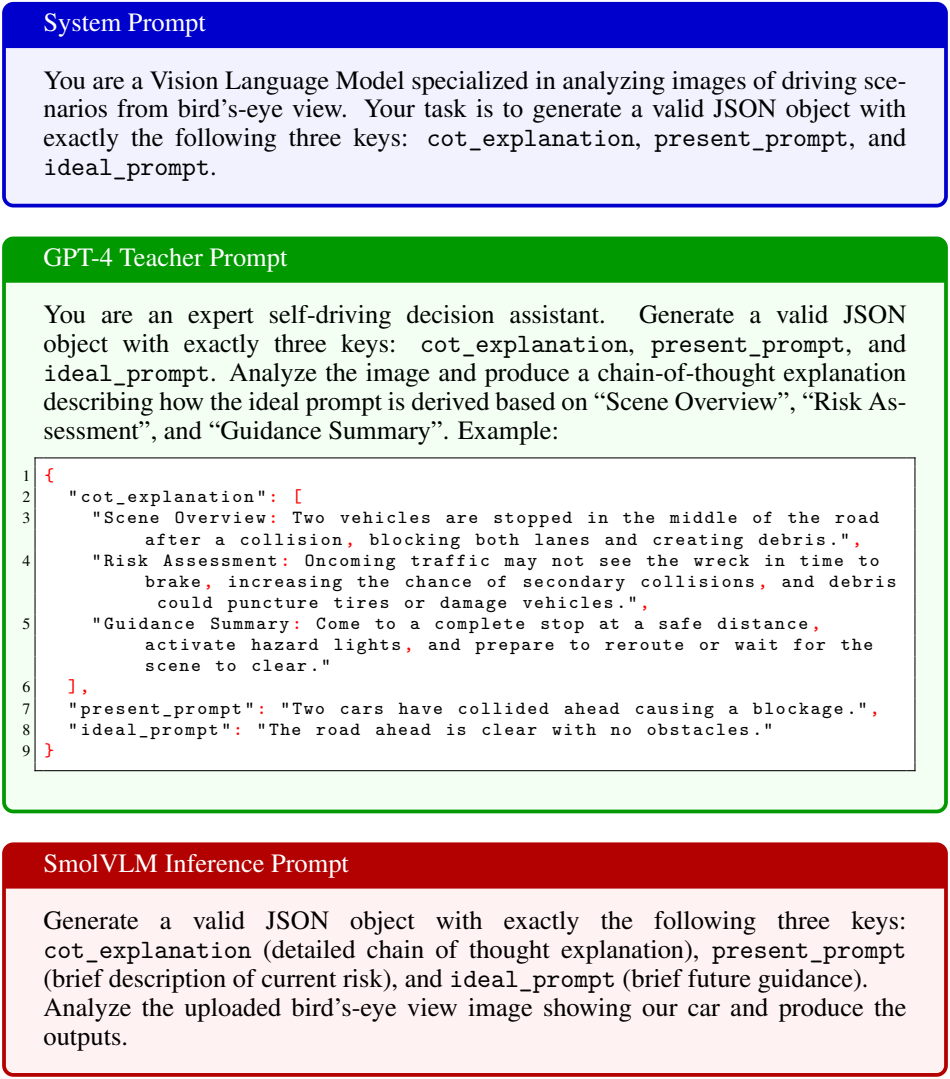

    \centering

    \begin{subfigure}{0.9\textwidth}
    \centering
    \begin{tcolorbox}[colframe=blue!80!black, colback=blue!5, title=System Prompt]
    You are a Vision Language Model specialized in analyzing images of driving scenarios from bird's-eye view. 
    Your task is to generate a valid JSON object with exactly the following three keys: \texttt{cot\_explanation}, \texttt{present\_prompt}, and \texttt{ideal\_prompt}.
    \end{tcolorbox}
    \end{subfigure}

    \vspace{1em}

\begin{subfigure}{0.9\textwidth}
\centering
\begin{tcolorbox}[colframe=green!60!black, colback=green!5, title=GPT-4 Teacher Prompt]
You are an expert self-driving decision assistant. Generate a valid JSON object with exactly three keys: \texttt{cot\_explanation}, \texttt{present\_prompt}, and \texttt{ideal\_prompt}. 
Analyze the image and produce a chain-of-thought explanation describing how the ideal prompt is derived based on ``Scene Overview'', ``Risk Assessment'', and ``Guidance Summary''. 
Example:

\begin{lstlisting}[language=json, basicstyle=\ttfamily\scriptsize, breaklines=true]
{
  "cot_explanation": [
    "Scene Overview: Two vehicles are stopped in the middle of the road after a collision, blocking both lanes and creating debris.",
    "Risk Assessment: Oncoming traffic may not see the wreck in time to brake, increasing the chance of secondary collisions, and debris could puncture tires or damage vehicles.",
    "Guidance Summary: Come to a complete stop at a safe distance, activate hazard lights, and prepare to reroute or wait for the scene to clear."
  ],
  "present_prompt": "Two cars have collided ahead causing a blockage.",
  "ideal_prompt": "The road ahead is clear with no obstacles."
}
\end{lstlisting}

\end{tcolorbox}
\end{subfigure}

    \vspace{1em}

    \begin{subfigure}{0.9\textwidth}
    \centering
    \begin{tcolorbox}[colframe=red!70!black, colback=red!5, title=SmolVLM Inference Prompt]
    Generate a valid JSON object with exactly the following three keys:
    \texttt{cot\_explanation} (detailed chain of thought explanation), 
    \texttt{present\_prompt} (brief description of current risk), 
    and \texttt{ideal\_prompt} (brief future guidance).
    
    Analyze the uploaded bird's-eye view image showing our car and produce the outputs.
    \end{tcolorbox}
    \end{subfigure}

    \caption{System, Training, and Inference Prompts used for Chain-of-Thought Distillation and Dynamic VLM Fine-Tuning.}
    \label{fig:prompt-examples}
\end{figure}

\paragraph{Chain‐of‐Thought Distillation and Dynamic VLM Fine‐Tuning.}
The dynamic VLM, SmolVLM‐256M, is fine‐tuned to generate “present” and “ideal” prompts along with chain‐of‐thought (CoT) rationales based on novelty-triggered events. For each BEV image tensor captured during driving, a GPT-4o-mini API as a teacher is queried with a fixed structured prompt to generate CoT outputs, including a risk assessment, a current scene description, and future guidance, serialized in JSON format. The agent processes single samples on-the-fly: each incoming image is paired with the teacher's labels to create a chat-formatted input that is used for lightweight adaptation of SmolVLM using Low-Rank Adaptation (LoRA). 

Fine-tuning operates asynchronously with a per-step update: one AdamW optimizer step with learning rate $5\times10^{-5}$, training a set of LoRA modules applied to the attention and MLP projections within the decoder. Only the LoRA parameters are updated; the base model weights and BEV image encoder remain frozen. No KL divergence regularization is used; instead, the student directly learns to reconstruct the full CoT and target prompts via cross-entropy loss on token predictions. Mini-batch training uses a batch size of 1 without gradient accumulation. Mixed-precision (FP16) training is enabled to accelerate updates and reduce memory overhead. After each fine-tuning step, LoRA adapter weights are saved for future inference. 

Dynamic VLM inference queries the most recent fine-tuned checkpoint when task novelty is detected ($\Delta_t > \delta$), generating prompts with a maximum of 256 output tokens. Generated JSON outputs are parsed to extract present prompts, ideal prompts, and chain-of-thought explanations, which are cached for downstream policy conditioning. All distillation and fine-tuning operations run asynchronously relative to the main control loop to minimize system latency. The prompts used to generate these outputs are displayed in Figure \ref{fig:prompt-examples}.

\begin{figure}[htbp]
  \centering
  \doublebox{%
    \begin{minipage}{0.95\textwidth}
      \textbf{Example 1: Tailgating on the Highway}\\[4pt]
      \textbf{Scene Overview:} You’re driving on the highway when a silver sedan closes in to within one car length behind you at high speed.\\
      \textbf{Risk Assessment:} The dangerously short following distance leaves virtually no reaction time if you brake, risking a rear-end collision.\\
      \textbf{Guidance Summary:} Gradually reduce speed to restore a two-second gap; signal and change lanes when safe.\\[4pt]
       \textbf{Present Prompt:} “Close tailgater risks rear-end crash.”\\
      \textbf{Ideal Prompt:} “Slow gently and maintain safe gap.”  
    \end{minipage}
  }

  \vspace{1em}

  \doublebox{%
    \begin{minipage}{0.95\textwidth}
      \textbf{Example 2: Sudden Braking of Lead Car}\\[4pt]
      \textbf{Scene Overview:} The red hatchback ahead flashes its brake lights and decelerates sharply without any visible obstacle.\\
      \textbf{Risk Assessment:} Unanticipated hard braking can catch you off-guard, leading to a collision if you cannot react in time.\\
      \textbf{Guidance Summary:} Immediately lift off the throttle, apply smooth braking to match their speed, and keep a safe buffer.\\[4pt]
      \textbf{Present Prompt:} “Car ahead brakes suddenly, risking crash.”\\
      \textbf{Ideal Prompt:} “Brake smoothly to keep safe distance.” 
    \end{minipage}
  }

  \vspace{1em}

  \doublebox{%
    \begin{minipage}{0.95\textwidth}
      \textbf{Example 3: Oncoming Car Drifting Across Centerline}\\[4pt]
      \textbf{Scene Overview:} Rounding a blind curve, an oncoming sedan drifts over the centerline into your lane.\\
      \textbf{Risk Assessment:} Encroachment at speed leaves little room to maneuver, risking a head-on collision.\\
      \textbf{Guidance Summary:} Decelerate sharply, steer toward the shoulder to maximize separation, and sound the horn to warn.\\[4pt]
      \textbf{Present Prompt:} “Oncoming car drifts over centerline, crash risk.”\\
      \textbf{Ideal Prompt:} “Brake and steer to safe shoulder.”  
    \end{minipage}
  }

    \vspace{1em}
  \doublebox{%
    \begin{minipage}{0.95\textwidth}
      \textbf{Example 4: Sudden Merge from Side Road}\\[4pt]
      \textbf{Scene Overview:} Exiting a parking lot on your right, a blue hatchback accelerates into your lane without signaling.\\
      \textbf{Risk Assessment:} Unexpected merger reduces your safe following distance and may force abrupt braking, risking a collision.\\
      \textbf{Guidance Summary:} Immediately lift off the throttle, apply controlled braking to increase gap, and prepare to steer away if necessary.\\[4pt]
      \textbf{Present Prompt:} “Merging car enters lane, crash hazard.”\\
      \textbf{Ideal Prompt:} “Slow down to safe following distance.”  
    \end{minipage}
  }

  \caption{Three driving scenarios illustrating how chain-of-thought prompting structures semantic reward generation. Each box shows the CoT breakdown which includes: Scene Overview, Risk Assessment, and Guidance Summary, followed by the negative “present Prompt” and positive “ideal Prompt” used to compute the Adaptive ideal-State Contrastive Reward.}
  \label{fig:cot_qualitative}
\end{figure}

\paragraph{Qualitative Impact of Chain-of-Thought Prompting.}  
Chain-of-thought prompting enhances reward quality by decomposing complex visual scenes into structured reasoning steps, which in turn guide the VLM to generate concise, context-specific prompts, as shown by examples in Figure \ref{fig:cot_qualitative}. In the tailgating example, a naïve prompt might merely state ``too close,'' but the CoT breakdown identifies speed and reaction time issues, yielding a ``present'' prompt that explicitly references high speed and collision risk, and an ``ideal'' prompt that prescribes restoring a two-second gap. This level of detail produces a semantic reward that penalizes the precise unsafe behavior rather than any generic proximity, leading to more informative gradients during training.

Similarly, in the sudden-braking and centerline-drift scenarios, CoT ensures that the VLM captures both hazard severity and corrective actions. The Guidance Summary step pinpoints the appropriate maneuver e.g. smooth matched braking or emergency lane offset, so that the “ideal” prompt encodes a concrete, physically realizable directive. This contrasts with single-shot prompts that often conflate multiple risks or issue vague advice. By aligning the semantic embedding with well-defined safety concepts and corresponding actions, chain-of-thought prompting yields denser, more accurate reward signals, reduces spurious correlations, and ultimately drives the agent toward safer, more reliable driving behaviors.


\paragraph{Compute Infrastructure}  
\begin{table}[ht]
  \centering
  \caption{\centering Compute Infrastructure}
  \label{tab:compute_infra}
  \begin{tabular}{lll}
    \toprule
    \textbf{Component}   & \textbf{Specification}                                           \\
    \midrule
    \emph{CPU}           & AMD EPYC 7413 (2 $\times$ 24 cores, 2 threads/core; 2 × 256 MiB L3)     \\
    \emph{System RAM}    & 128 GB DDR4                                                    \\
    \emph{GPUs}          & 2 $\times$ NVIDIA RTX 3070 (24 GB GDDR6)  \\
    \emph{Storage}       & NVMe SSD (Samsung PM983, 1 TB)                                  \\
    \emph{Network NIC}   & Intel X550T 10 GbE (2 ports)                                    \\
    \emph{OS}            & Ubuntu 20.04.6 LTS, Kernel 5.4.0-172                            \\
    \emph{Frameworks}    & PyTorch 2.0, CUDA 11.7, NCCL                                    \\
    \emph{RL Library}    & Stable-Baselines3                                               \\
    \emph{Other Tools}   & OpenAI CLIP and GPT-4, Hugging Face SmolVLM                     \\
    \bottomrule
  \end{tabular}
\end{table}

As summarized in Table~\ref{tab:compute_infra}, all experiments were conducted on a dual‐socket AMD EPYC 7413 server, with each socket providing 24 physical cores (48 hardware threads) clocked up to 2.65 GHz and backed by 128 GB of DDR4‐3200 memory in a quad‐channel configuration.  The system’s PCIe 4.0 lanes deliver high‐bandwidth connectivity to a pair of NVIDIA RTX 3070 GPUs, each equipped with 24 GB of GDDR6 memory.  One GPU was dedicated to hosting the CARLA simulation server (including sensor rendering and traffic management), while the second handled all reinforcement learning workloads, world‐model ensemble training, and on-demand VLM prompt generation.  The host runs Ubuntu 20.04.6 LTS (kernel 5.4.0-172), NVIDIA driver 470.239.06, and CUDA 11.4, with our deep learning stack built on Python 3.10, PyTorch 2.0 (using the CUDA 11.7 runtime), and NCCL for multi-GPU communication.  Reinforcement learning algorithms (SAC and PPO) are implemented via the Stable-Baselines3 library, with mixed‐precision support provided by Apex to accelerate training.  Semantic reasoning leverages a frozen CLIP ViT-bigG-14 model alongside a 256 M-parameter SmolVLM fine-tuned via chain-of-thought distillation using GPT-4, both served from host RAM to minimize I/O overhead. Under these conditions, a full $10^6$-step \texttt{DriveMind} training run (including policy updates, world-model optimization, and novelty-triggered VLM decoding) requires approximately 56 wall-clock hours when distributed across the two GPUs.  All baseline methods were evaluated on the identical hardware and software configuration to ensure a fair and reproducible comparison.

\begin{figure}
  \centering
  \includegraphics[width=\textwidth]{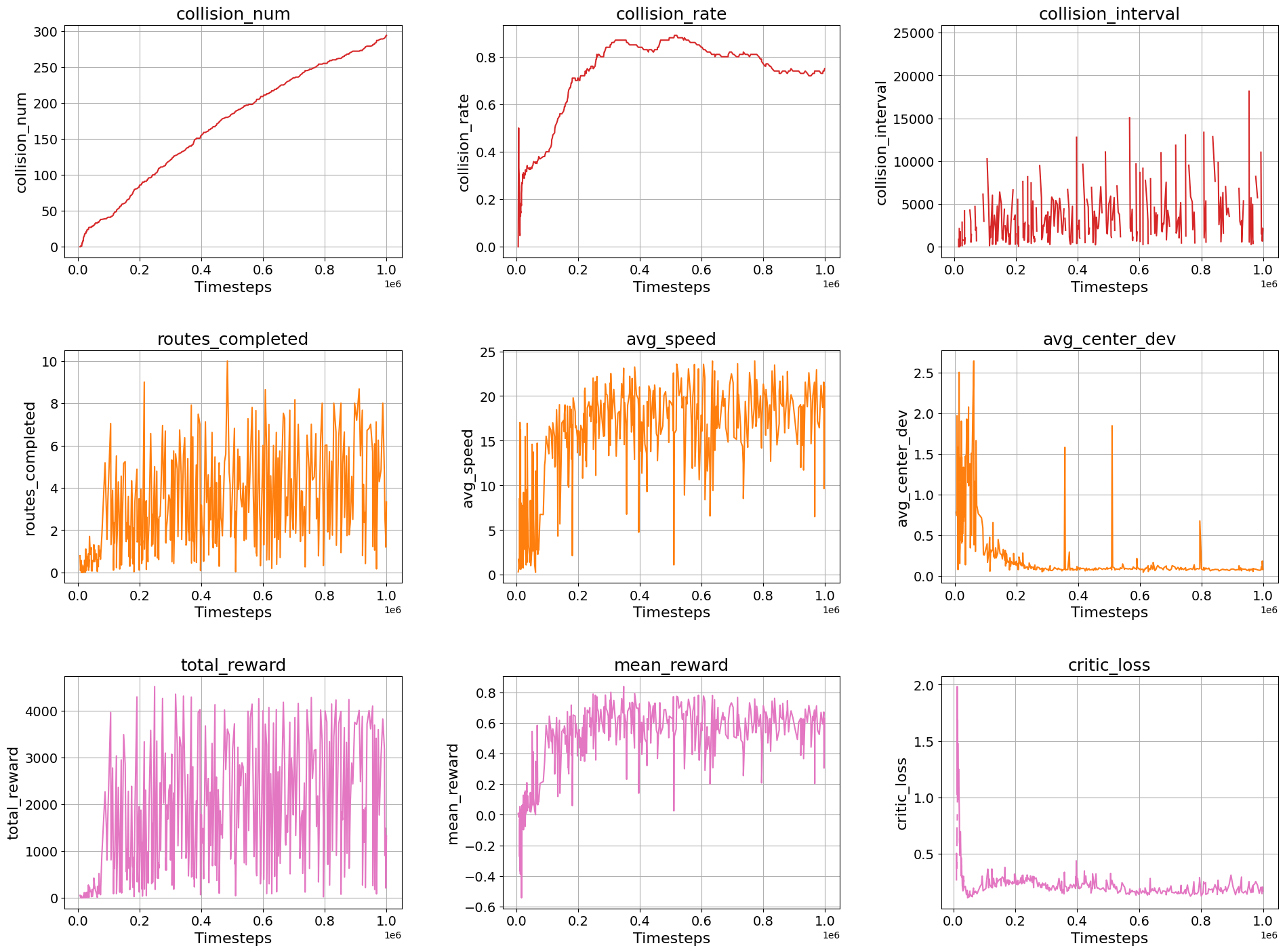}
  \caption{
    \textbf{\texttt{DriveMind} Training Dynamics over 1M Timesteps.}
    \emph{Top row (red):} Collision statistics, the cumulative number of collisions (\textit{collision\_num}), instantaneous collision rate (\textit{collision\_rate}), and the average interval between collisions (\textit{collision\_interval}). 
    \emph{Middle row (orange):} Driving performance metrics, number of completed routes per episode (\textit{routes\_completed}), average speed (\textit{avg\_speed}), and mean lateral deviation from lane center (\textit{avg\_center\_dev}). 
    \emph{Bottom row (pink):} Learning signals, the total episodic reward (\textit{total\_reward}), mean per-step reward (\textit{mean\_reward}), and the SAC critic’s Bellman-error loss (\textit{critic\_loss}). 
  }
  \label{fig:drivemind_training_curves}
\end{figure}

\subsection{Training Results}

We now examine \texttt{DriveMind}’s learning dynamics over the course of training. Figure~\ref{fig:drivemind_training_curves} presents nine key metrics plotted against the number of environment steps (up to $1\times10^{6}$). These curves collectively illustrate how the agent transitions from an untrained policy (which maybe prone to frequent crashes and erratic control) to a stable, high-performance driving strategy.

In the top row (red curves), we track collision statistics. The leftmost panel shows the cumulative number of collisions climbing rapidly during early exploration and then flattening as the agent acquires avoidance behavior. The middle plot, collision rate, peaks near 0.9 in the first $2\times10^{5}$ steps before steadily declining to around 0.75, reflecting fewer crashes per timestep. The rightmost panel, collision interval, measures the average steps between collisions; it expands from the low hundreds into the tens of thousands, signifying that unsafe events become both rarer and more widely spaced.

The middle row (orange curves) depicts driving efficacy. Routes completed per episode start below one and rise into the 5, 8 range by the end of training, demonstrating improved mission success. Average forward speed increases from near zero to a stable 15, 20 km/h band, while average lateral deviation collapses from over two meters to near zero, confirming the agent learns to stay centered in its lane.

Finally, the bottom row (pink curves) shows learning signals. The total reward per episode grows from scattered low values to a plateau near 3,000-4,000, indicating the agent consistently earns positive reinforcement. The mean per-step reward rises from negative to about 0.6, demonstrating that most actions become beneficial. Critic loss falls sharply from around 2.0 to 0.2 within the first 100,000 steps and remains low, evidencing stable Q-function convergence.

\begin{table}[ht]
\centering
\caption{Per‐step latency breakdown on dual RTX 3070 (in milliseconds, two decimal places).}
\label{tab:latency_breakdown}
\begin{tabular}{lc}
\toprule
\textbf{Component}                                  & \textbf{Latency (ms)} \\
\midrule
SAC policy inference \& action selection           & 3.87 \\
Contrastive VLM embedding                           & 1.58 \\
Novelty detection                                   & 2.97 \\
Prompt‐cache lookup                                 & 0.11 \\
World‐model predictive foresight                    & 0.53 \\
\midrule
Baseline per‐step total (no dynamic VLM)            & 9.06 \\
Dynamic VLM prompt generation (when triggered)      & 2975.23 \\
Amortized dynamic VLM cost ($\sim$1 per 100 steps)       & 29.75 \\
\midrule
Overall per‐step latency (amortized)                & 38.81 \\
\bottomrule
\end{tabular}
\end{table}

\subsection{Runtime Latency for Assessing Real-Time Deployability} \label{appen:real-time-deloyability}
We measure per-step latency by averaging over 1,000 consecutive control loop iterations during a typical CARLA Town 2 rollout, capturing end-to-end timing from observation to action. Table~\ref{tab:latency_breakdown} reports the mean wall-clock times for each component on our dual RTX 3070 setup, with two decimal-place precision. SAC policy inference and action selection incur only 3.87 ms per step, reflecting the efficiency of our lightweight actor network. Contrastive VLM embedding adds 1.58 ms, while novelty detection and prompt-cache lookup contribute 2.97 ms and 0.11 ms respectively. The world-model predictive foresight module runs in 0.53 ms. Summing these yields a baseline per-step latency of 9.06 ms without dynamic prompting. When the dynamic VLM is triggered (on average once every 100 steps), prompt generation takes 2975.23 ms; amortized over 100 steps, this adds 29.75 ms. Thus, the overall amortized per-step latency is 38.81 ms, corresponding to approximately 25 Hz control. These results confirm that \texttt{DriveMind} can support near real-time operation, with dynamic semantic prompting (on-demand) integrated without compromising the responsiveness required for safe autonomous driving.

\end{document}